\documentclass[sigconf]{acmart}

\usepackage{amsmath}
\usepackage{amsthm}
\usepackage{balance}
\usepackage[capitalize,noabbrev]{cleveref}
\usepackage{courier}
\usepackage{enumitem}
\usepackage{graphicx}
\usepackage{hyperref}
\usepackage{mathtools}
\usepackage{multirow}
\usepackage{listings}
\usepackage{ragged2e}
\usepackage{threeparttable}
\usepackage{tcolorbox}
\usepackage[table]{xcolor}
\usepackage{xspace}
\usepackage{xcolor}

\definecolor{codegreen}{rgb}{0,0.45,0}
\definecolor{codegray}{rgb}{0.45,0.45,0.45}
\definecolor{codepurple}{rgb}{0.50,0,0.50}
\definecolor{backcolour}{rgb}{0.98,0.98,0.98}
\AtBeginDocument{%
  }

\copyrightyear{2026}
\acmYear{2026}
\setcopyright{cc}
\setcctype{by}
\acmConference[KDD '26]{Proceedings of the 32nd ACM SIGKDD Conference on Knowledge Discovery and Data Mining V.2}{August 09--13, 2026}{Jeju Island, Republic of Korea}
\acmBooktitle{Proceedings of the 32nd ACM SIGKDD Conference on Knowledge Discovery and Data Mining V.2 (KDD '26), August 09--13, 2026, Jeju Island, Republic of Korea}
\acmDOI{10.1145/3770855.3818846}
\acmISBN{979-8-4007-2259-2/2026/08}

\begin{document}

\title{Enhancing Spatial Reasoning in Large Language Models for\\ Metal-Organic Frameworks Structure Prediction}

\author{Mianzhi Pan}
\authornote{Both authors contributed equally to this research.}
\affiliation{%
  \institution{National Key Laboratory for Novel Software Technology\\Nanjing University}
  \city{Nanjing}
  \country{China}
}
\email{panmz@smail.nju.edu.cn}

\author{Jianfei Li}
\authornotemark[1]
\affiliation{%
  \institution{National Key Laboratory for Novel Software Technology\\Nanjing University}
  \city{Nanjing}
  \country{China}
}
\email{lijf@smail.nju.edu.cn}

\author{Peishuo Liu}
\affiliation{%
  \institution{National Key Laboratory for Novel Software Technology\\Nanjing University}
  \city{Nanjing}
  \country{China}
}
\email{liupeishuo@smail.nju.edu.cn}

\author{Botian Wang}
\affiliation{%
  \institution{Institute of AI Industry Research (AIR)\\ Tsinghua University}
  \city{Beijing}
  \country{China}
}
\email{wbt23@mails.tsinghua.edu.cn}

\author{Yawen Ouyang}
\affiliation{%
 \institution{Shanghai Artificial Intelligence Laboratory}
 \city{Shanghai}
 \country{China}}
\email{ouyangyawen@air.tsinghua.edu.cn}

\author{Yiming Rong}
\affiliation{%
  \institution{University of Chinese Academy of Sciences}
  \city{Beijing}
  \country{China}}
\email{rongyiming2022@ia.ac.cn}

\author{Wei-Ying Ma}
\affiliation{%
  \institution{Institute of AI Industry Research (AIR)\\ Tsinghua University}
  \city{Beijing}
  \country{China}
}
\email{maweiying@air.tsinghua.edu.cn}

\author{Hao Zhou}
\authornote{Corresponding author.}
\affiliation{%
  \institution{Institute of AI Industry Research (AIR)\\ Tsinghua University}
  \city{Beijing}
  \country{China}
}
\email{zhouhao@air.tsinghua.edu.cn}

\author{Jianbing Zhang}
%\authornote{Corresponding author.}
\authornotemark[2]
\affiliation{%
  \institution{National Key Laboratory for Novel Software Technology \& ChemBIC\\Nanjing University}
  \city{Nanjing}
  \country{China}
}
\email{zjb@nju.edu.cn}

\renewcommand{\shortauthors}{Mianzhi Pan et al.}
%% No italics, no superscripts, not anonymous
%% Use footnote or author note to identify equal contribution, shared contribution, and/or contact author info
\newcommand{\methodname}{MOF-LLM\xspace}

\begin{abstract}
Metal-organic frameworks (MOFs) are porous crystalline materials with broad applications such as carbon capture and drug delivery, yet accurately predicting their 3D structures remains a significant challenge.
While Large Language Models (LLMs) have shown promise in generating crystal structures, their application to MOFs is hindered by MOFs' high structural complexity arising from the large number of atoms in unit cell.
Inspired by the success of block-wise paradigms in deep generative models for MOFs, we pioneer the application of LLMs in this domain by introducing \methodname, the first LLM framework specifically adapted for block-level MOF structure prediction. 
To effectively harness LLMs for this 3D modular assembly task, our training paradigm integrates spatial-aware continual pre-training (CPT), structural supervised fine-tuning (SFT), and matching-driven reinforcement learning (RL). By incorporating explicit spatial priors and optimizing structural stability via Soft Adaptive Policy Optimization (SAPO), our approach substantially enhances the spatial reasoning in a Qwen-3 8B model for MOF structure prediction. 
Comprehensive experiments demonstrate that \methodname achieves state-of-the-art performance with a match rate of 35.78\% while exhibiting superior sampling efficiency of 0.04 seconds per structure.\footnote{The model, data and code are available at \url{https://github.com/panmianzhi/MOF-LLM}.}
% Comprehensive experiments demonstrate that \methodname outperforms state-of-the-art denoising-based and LLM-based methods while exhibiting superior sampling efficiency.
\end{abstract}




\maketitle

\section{Introduction}
\begin{figure}[t]
  \includegraphics[width=0.97\linewidth]{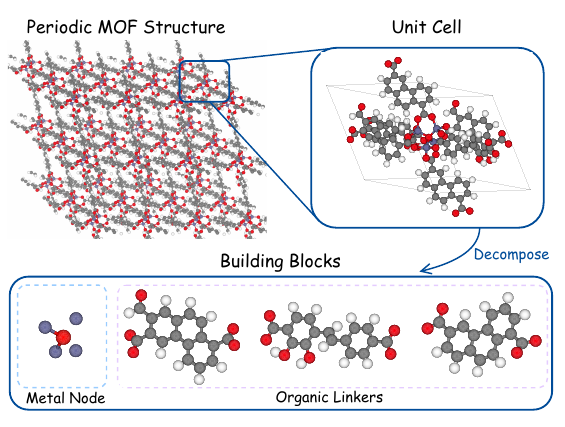}
  \caption{Illustration of MOF structures. An MOF structure and its unit cell are shown on top. The unit cell can be decomposed into building blocks, including metal nodes and organic linkers (bottom). Atom color: Zn (purple), O (red), C (gray), H (white).}
  \label{fig:intro}
\end{figure}

% introduce mof
% Metal-organic frameworks (MOFs) are a revolutionary class of porous crystalline materials with extensive applications such as carbon capture~\citep{mcdonald2015cooperative}, drug delivery~\citep{horcajada2010porous}, and water harvesting~\citep{kim2017water}. In light of their value, the development of MOFs was recently awarded the Nobel Prize in Chemistry~\citep {nobel2025chemistry}. 
% Structurally, MOFs comprise inorganic metal nodes and organic linkers, i.e., building blocks, assembling into periodic reticular architectures.
% Variations in the choice of building blocks lead to millions of MOF structures with diverse properties~\citep{moosavi2020understanding}. Consequently, navigating this expansive design space for tailored functionalities has created a growing demand for rapid and accurate 
% prediction of MOF structures.
Metal-organic frameworks (MOFs) are a revolutionary class of porous crystalline materials with extensive applications such as carbon capture~\citep{mcdonald2015cooperative}, drug delivery~\citep{horcajada2010porous}, and water harvesting~\citep{kim2017water}. In light of their value, the development of MOFs was recently awarded the Nobel Prize in Chemistry~\citep {nobel2025chemistry}.
The functional versatility of MOFs arises from their distinctive architecture, where inorganic metal nodes and organic linkers, i.e., building blocks, assemble into periodic reticular structures (\Cref{fig:intro}).
Variations in these building blocks can lead to millions of possible structures with diverse properties~\citep{moosavi2020understanding}.
% Therefore, navigating this expansive design space necessitates rapid and accurate structural characterization.
% Crystal structure prediction (CSP), which aims to identify the thermodynamically stable arrangement for a given chemical composition, plays a central role in accelerating functional MOF discovery.
% Botian revision
Navigating this expansive space to identify optimal candidates poses a significant challenge, as traditional experimental trial-and-error is intractable~\cite{boyd2019data}.
Therefore, Crystal Structure Prediction (CSP) plays a pivotal role in accelerating MOF discovery. By computationally predicting thermodynamically stable 3D structures, CSP enables the rapid screening of materials prior to laboratory synthesis.

Recently, Large Language Models (LLMs) have emerged as a promising paradigm for CSP. By leveraging extensive pre-trained knowledge, LLMs have demonstrated a strong ability to generate physically plausible crystal structures when fine-tuned on crystallographic text data~\citep{gruverfine,antunes2024crystal,sriram2024flowllm,xu2025plaid++,wang2025crystalicl}. 
Such crystal representations typically consist of atom-level information, including element types and atomic positions.
However, extending these approaches to MOFs presents significant challenges. Unlike simple bulk materials, MOF unit cells often contain hundreds of atoms. Representing such massive systems as conventional atom-wise strings leads to \textbf{excessively long contexts} that current LLMs struggle to handle.

Inspired by recent advances in deep generative models for CSP, we adopt a \textbf{block-wise} generation paradigm to address the scalability challenges in MOF structure prediction. While early CSP models operated at the atomistic level~\citep{jiao2023crystal,miller2024flowmm,wu2025periodic}, recent MOF-specific approaches have achieved significant breakthroughs by exploiting the intrinsic modularity of MOFs~\citep{fumofdiff,kimmofflow,jiao2025mof}. 
These methods predict the positions and orientations of MOF building blocks rather than individual atomic coordinates. 
As a result, the block-level generation paradigm substantially reduces the dimensionality of the design space. 
% Building on this idea, we adapt the block-wise paradigm and reformulate MOF structure prediction as an LLM-driven `LEGO building' task at the microscale.

Despite their promise, LLMs face key challenges in block-wise MOF structure prediction.
First, MOF building blocks possess \textbf{complex 3D geometries} that are difficult for LLMs to understand, frequently resulting in block collisions and physically implausible assemblies.
Second, generating stable MOFs requires \textbf{precise control} over the orientations of individual blocks, necessitating the accurate generation of rotations in 3D space.
However, existing LLMs exhibit \textit{limited spatial reasoning ability}, particularly regarding rotations~\citep{lv2025atomworld}, thereby constraining their efficacy in this task.

To address these challenges, this paper proposes \methodname, the first LLM training framework that integrates \textit{spatial-aware continue pre-training} (CPT), \textit{structural supervised fine-tuning} (SFT), and \textit{matching-driven reinforcement learning} (RL) for autoregressive MOF structure prediction. Our pipeline is as follows:
first, we perform CPT on our curated dataset that explicitly incorporates block connectivity and geometry. This process facilitates LLMs' basic spatial understanding and reasoning for MOF structures.
Subsequently, we conduct SFT to generate lattice parameters and block-wise roto-translations, thereby enabling LLMs to assemble building blocks.
Finally, we adapt RL with reward signals from the matching degree between the generated structures and stable MOF references, which further improves the stability of the predicted structures.

To validate the effectiveness of \methodname, we conducted comprehensive experiments on the MOF dataset introduced by \citet{boyd2019data}.
\methodname demonstrates \textbf{superior predictive performance}, achieving a match rate of \textbf{35.78\%} on unseen MOFs and outperforming both existing LLM-based approaches and state-of-the-art denoising-based models.
% \methodname demonstrates \textbf{state-of-the-art predictive performance}, achieving a match rate of \textbf{35.78\%} on unseen MOFs and outperforming both existing LLM-based approaches and denoising-based models.
Beyond strong accuracy, \methodname also exhibits \textbf{exceptional computational efficiency}: with an inference time of \textbf{0.04} seconds per structure, it exceeds the fastest denoising-based methods currently available. 

Our main contributions are as follows:
\begin{itemize}[leftmargin=*,itemsep=3pt,topsep=1pt,parsep=0pt]
    \item We propose \methodname, a progressive framework integrating spatial-aware CPT, structural SFT, and matching-driven RL. To our knowledge, this is the first work to adopt LLMs for MOF structure prediction.
    \item We enhance the spatial reasoning ability of LLMs, allowing accurate assembly of 3D building blocks into stable MOF structures.
    \item We demonstrate the superior performance of \methodname over all baselines in MOF structure prediction while achieving exceptional computational efficiency. Our analysis reveals the effectiveness of each training stage.
\end{itemize}

\section{Related Work}
\textbf{Generative Models for Crystal Structure Prediction (CSP).} 
Deep generative models are promising solutions for CSP. \citet{jiao2023crystal,jiaospace} propose joint generation of lattice parameters and fractional atomic coordinates using a periodic \textit{E}(3)-equivariant denoising diffusion model. \citet{miller2024flowmm} and \citet{wu2025periodic} introduce Riemannian flow matching (RFM) and periodic Bayesian flow networks (BFNs), respectively, enabling efficient sampling of crystals. 
For MOFs, \citet{kimmofflow} treat the metal nodes and organic linkers as rigid building blocks and introduce an RFM model to predict the position and orientations of these components. 
\citet{jiao2025mof} extend the periodic BFNs to the hypersphere for better orientation modeling by introducing Bingham BFNs. 
Unlike these denoising-based paradigms, we pioneer the application of LLMs for autoregressive MOF structure prediction.

\textbf{LLM-based Crystal Generation.} Although denoising-based models currently dominate 3D structure generation, recent advances have adapted LLMs for crystal generation by treating structures as text sequences. \citet{gruverfine} and \citet{antunes2024crystal} pioneer fine-tuning LLMs to produce text-encoded atomistic data for crystals. Furthermore, CrystalICL~\citep{wang2025crystalicl} extends LLM-based crystal generation with hybrid instruction tuning to enable few-shot in-context learning. \citet{xu2025plaid++} fine-tunes a Qwen-2.5 backbone~\citep{qwen2025qwen25technicalreport} with reinforcement learning to bias generation towards stable and novel structures with desired properties. MatExpert~\citep{ding2025matexpert} mimics the workflow of human material scientists to generate crystals within a retrieval–transition–generation framework. 
To improve structural precision, FlowLLM~\citep{sriram2024flowllm} proposes a hybrid framework that couples LLMs with an RFM model~\citep{miller2024flowmm}, combining the ability of LLMs to provide informative structural priors with the continuous optimization strength of denoising-based models.
Despite these advances, existing atom-level text representations fail to capture the intrinsic topology and modularity of MOFs, and also lead to extremely long contexts that LLMs struggle with.

% Botian addition: 
Specifically for MOFs, recent systems like MOFGPT~\citep{badrinarayanan2025mofgpt} and MOFGen~\citep{inizan2025system} leverage LLMs for generative design. However, existing approaches circumvent direct 3D structural generation. They either rely on external diffusion solvers~\citep{inizan2025system} or simplify topologies into 1D string identifiers~\citep{badrinarayanan2025mofgpt}. Consequently, the potential of LLMs to explicitly perceive 3D spatial relationships and directly assemble precise atomic structures remains largely untapped.

\textbf{LLM for 3D Object Generation.} Emerging research has begun to explore the application of LLMs in generating physically plausible and realizable 3D objects. 
LLaMA-Mesh~\citep{wang2024llama} fine-tunes the LLaMA model~\citep{dubey2024llama} to directly output 3D meshes, enabling unified language-mesh generation and spatial understanding within a single model. Similarly, BrickGPT~\citep{pun2025generating} fine-tunes an LLM on a carefully-curated brick assembly dataset to generate physically stable and constructible brick structures from textual descriptions.
These methods focus on positioning geometric primitives. In contrast, we focus on assembling intricate MOF blocks with full 3D roto-translational freedom.

\section{Preliminaries}
\subsection{MOF Representation}\label{sec:mof_repr}
\textbf{Atom-level Representation of MOFs.} Generally, MOFs are crystalline materials composed of atoms arranged periodically in 3D space. Such periodicity is described by a \emph{unit cell}, the smallest repeating structural unit whose translations generate the infinite crystal.
A unit cell containing $N$ atoms can be represented by the triplet $\mathcal{S}=(\mathbf{A}, \mathbf{X}, \mathbf{L})$. Here, $\mathbf{A}=[a_n]_{n=1}^N \in \mathbb{A}^N$ with $\mathbb{A}$ specifies the atomic species, $\mathbf{X}=[\mathbf{x}_n]_{n=1}^N \in \mathbb{R}^{N\times 3}$ gives the corresponding Cartesian coordinates of the atoms, and $\mathbf{L}=[\mathbf{l}_1,\mathbf{l}_2,\mathbf{l}_3] \in \mathbb{R}^{3\times 3}$ is the lattice matrix that encodes the periodicity of the crystal, which is uniquely determined by six lattice parameters: the edge lengths $(a,b,c)\in\mathbb{R}_{+}^3$ and the interaxial angles $(\alpha,\beta,\gamma)\in[0,180]^3$.
The infinite structure is obtained by 
\begin{equation}
   \left\{(a_n', \mathbf{x}_n') \mid a_n' = a_n,\; \mathbf{x}_n' = \mathbf{x}_n + \mathbf{k}\mathbf{L},\; \mathbf{k} \in \mathbb{Z}^{1\times 3}\right\},
\end{equation}
where $\mathbf{k}$ is an integral vector that translates the unit cell to tile the entire space.

In crystallography, the fractional coordinates are usually used to reflect the periodicity of the crystal structure, utilizing the lattice vectors $[\mathbf{l}_1,\mathbf{l}_2,\mathbf{l}_3]$ as the basis vectors instead of the standard Cartesian basis. Accordingly, the Cartesian coordinates $\mathbf{X}$ can be replaced by fractional coordinates $\mathbf{F}=\mathbf{X}\mathbf{L}^{-1}\in [0,1)^{N\times 3}$. This representation is widely adopted by contemporary methods for crystal generation~\citep{jiao2023crystal,wu2025periodic,gruverfine,xu2025plaid++}. 

\textbf{Block-level Representation of MOFs.} Inherently, MOFs are modular assemblies of building blocks (metal nodes and organic linkers). 
This modularity has motivated several studies to adopt block-level representations of MOFs~\citep{kimmofflow,kim2025flexible,jiao2025mof}. 
Formally, a unit cell decomposed into $M$ blocks can be represented as the tuple $\mathcal{S}=(\mathcal{B}, \mathbf{T}, \mathbf{R}, \mathbf{L})$. 
Here, $\mathcal{B}=[\mathbf{B}_m]_{m=1}^M$ denotes the set of building blocks, and each building block $\mathbf{B}_m=(\mathbf{A}_m, \mathbf{X}_m)$ is a group of $N_m$ atoms with $\mathbf{A}_m\in\mathbb{A}^{N_m}$ specifies the atomic species and $\mathbf{X}_m\in\mathbb{R}^{N_m\times 3}$ denotes the \emph{local} atomic coordinates.
The local coordinate system is defined using principal component analysis (PCA)-based equivariant axes.
$\mathbf{T}\in\mathbb{R}^{M\times 3}$ and $\mathbf{R}\in SO(3)^M$ are translation vectors and rotation matrices, which specify the center position and orientation of each building block, respectively. See Appendix \ref{app:local_coord} for construction details of the local coordinate system and associated roto-translations.

\begin{figure*}[t]
  \includegraphics[width=\textwidth]{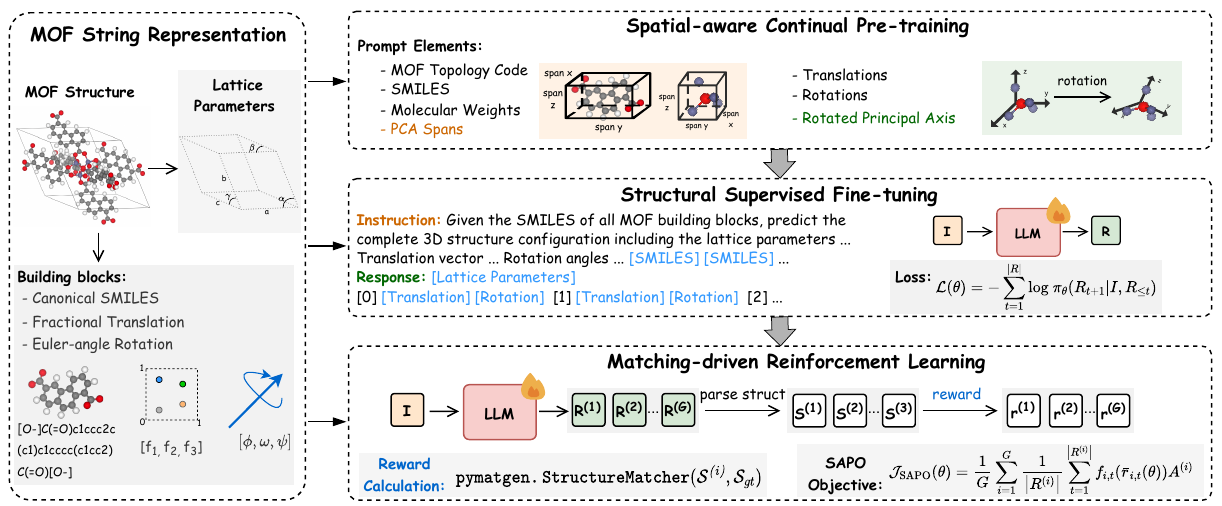}
  \caption{Overview of our \methodname framework. It enhances the spatial reasoning ability of LLMs for MOF structure prediction via a three-stage training: spatial-aware CPT to inject spatial and geometric priors, structural SFT enables block assembly, and matching-driven RL refines structure stability using Soft Adaptive Policy Optimization (SAPO).}
  \label{fig:method}
\end{figure*}

\subsection{MOF Structure Prediction}
Following established task settings~\citep{kimmofflow,jiao2025mof}, we employ the block-level representation and formulate MOF structure prediction as a generative assembly problem. Specifically, the objective is to predict the lattice matrix along with the roto-translations for a given set of MOF building blocks.
Formally, given a dataset $\mathcal{D}$ of stable MOFs, the training objective is to maximize the following likelihood:
\begin{equation}
\max_{\theta}\mathbb{E}_{(\mathcal{B}, \mathbf{T}, \mathbf{R}, \mathbf{L})\sim\mathcal{D}}\left[ p_\theta(\mathbf{T}, \mathbf{R}, \mathbf{L} \mid \mathcal{B}) \right].
\end{equation}
During inference, $\mathbf{T}$, $\mathbf{R}$, and $\mathbf{L}$ are sampled from $p_{\theta}(\mathbf{T}, \mathbf{R}, \mathbf{L} \mid \mathcal{B})$ to assemble the building blocks, thereby reconstructing atom-level MOF structures for final evaluation.

\section{Method}
We now introduce \methodname. First, in \Cref{method_1}, we explain how we convert the 3D MOF structures to plain text for easy processing by LLMs. 
\Cref{method_2} and \Cref{method_3} then detail the spatial-aware continual pre-training (CPT) and the block-assembly supervised fine-tuning (SFT) process, respectively, which together enable a pre-trained LLM (Qwen-3 8B)~\citep{yang2025qwen3} to accurately predict MOF structures. 
Finally, \Cref{method_4} presents the Soft Adaptive Policy Optimization (SAPO)~\citep{gao2025soft} with a matching-based reward that biases LLMs toward generating more stable structures. The overall framework is shown in \Cref{fig:method}.
 
\subsection{Text Formatting of MOF Structure}\label{method_1}
\textbf{Lattice and Translations.} Enabling LLMs to directly generate 3D MOF structures requires an effective textual representation of the structures $\mathcal{S}=(\mathcal{B}, \mathbf{T}, \mathbf{R}, \mathbf{L})$. Recent studies~\citep{antunes2024crystal,gruverfine} have demonstrated that fine-tuned LLMs can generate crystal structures encoded via lattice parameters $(a,b,c,\alpha,\beta,\gamma)$ and fractional atomic coordinates. Aligning with this approach, we convert the lattice matrix $\mathbf{L}$ to scalar lattice parameters (each rounded to two decimal places) and transform the translation vectors to fractional coordinates defined as $\mathbf{F}=\mathbf{T}\mathbf{L}^{-1}$ (rounded to three decimal places).

\textbf{Rotations.} Directly utilizing rotation matrices is challenging for LLMs to generate, as these representations are redundant, conceptually abstract, and strictly constrained by the orthogonality conditions of the Special Orthogonal group $SO(3)$.
To obtain a compact, unconstrained, and more intuitive representation, we adopt Euler angles $(\phi, \omega, \psi)$, where $\phi\in [-\pi, \pi]$ (roll), $\omega\in [-\pi/2, \pi/2]$ (pitch), and $\psi\in [-\pi,\pi]$ (yaw) are rotation angles that parameterize a 3D rotation $\mathbf{R}$ as three successive elemental rotations around the $x$-$y$-$z$ coordinate axes by
\begin{equation}
\begin{aligned}
\mathbf{R} =\ &\mathbf{R}_{z}(\psi) \mathbf{R}_{y}(\omega) \mathbf{R}_{x}(\phi) \\
=\
&\begin{bmatrix}
c_\psi c_\omega & c_\psi s_\omega s_\phi - s_\psi c_\phi & c_\psi s_\omega c_\phi + s_\psi s_\phi \\
s_\psi c_\omega & s_\psi s_\omega s_\phi + c_\psi c_\phi & s_\psi s_\omega c_\phi - c_\psi s_\phi \\
-s_\omega & c_\omega s_\phi & c_\omega c_\phi
\end{bmatrix}
\end{aligned}
\end{equation}
with $c_{(\cdot)}=\mathrm{cos}(\cdot)$ and $s_{(\cdot)}=\mathrm{sin}(\cdot)$. Appendix \ref{app:rotation} provides the derivation of Euler angles from the rotation matrix $\mathbf{R}$. The Euler angles are rounded to three decimal places in our experiment.

\textbf{Building Blocks.}
The building blocks $\mathcal{B}$ are complex 3D geometries that present a significant semantic gap for LLMs.
However, since these blocks correspond to chemically meaningful entities, i.e., metal ion clusters and organic molecules, we adopt canonical SMILES~\citep{weininger1989smiles,o2012towards} as a solution.
In contrast to basic SMILES algorithms, canonical SMILES ensures that it produces a unique string for any given molecule.
For an extracted molecular graph from 3D structures, the canonicalization algorithm systematically selects a specific initial atom and traverses the molecule using a predefined set of rules.
This representation naturally aligns with LLMs and enables exploitation of the chemistry understanding and semantic priors of SMILES, which are implicitly acquired during large-scale pre-training~\citep{guo2023can,castro2023large,wu2024leveraging}.
To compensate for the loss of explicit 3D geometric information resulting from converting structures to 1D strings, we introduce a spatially aware CPT process in the subsequent section.

\subsection{Spatial-aware Continual Pre-training}\label{method_2}
The assembly of building blocks into stable MOF structures can be regarded as `LEGO building' at the microscale.
A critical factor for success is the model's ability to perceive the intrinsic geometry of individual blocks and to reason about their roto-translations.
To this end, we augment the description of each building block $\mathbf{B}_m$ with its molecular weight and spatial spans $[s_m^1,s_m^2,s_m^3]$ along three PCA axes.
Specifically, the spans are defined as
\begin{equation}
    s_m^{\,i}
=
\max\left( \mathbf{X}_m \mathbf{p}_m^{i} \right)
-
\min\left( \mathbf{X}_m \mathbf{p}_m^{i} \right),\quad i=1,2,3
\end{equation}
where $\mathbf{X}_m$ is the local coordinates and $\mathbf{p}_m^{i}\in\mathbb{R}^3$ is the $i$-th PCA axis of block $\mathbf{B}_m$.
Together with the molecular weight, the PCA spans form a mass-weighted minimal bounding box that encloses the building block and provides a compact geometric descriptor.

To further elucidate the impact of rotation on block orientation, we explicitly provide the block's principal axis before and after the applied rotation.
In addition, we incorporate the topology code extracted by \texttt{MOFid}~\citep{bucior2019identification}, which describes the underlying connectivity of the building blocks and promotes a holistic understanding of the global MOF structure.
A \emph{simplified} prompt template is presented below for illustration:
\begin{tcolorbox}[
  colback=white,
  colframe=black!20,
  fontupper=\ttfamily\small,
  halign=justify,
  left=1.5mm,
  right=0mm
]
The MOF structure has the \textcolor{blue}{[Topo} \textcolor{blue}{Code]} topology, defined as the standard symbol in the Reticular Chemistry
Structure Resource (RCSR) database. \textcolor{blue}{[Topo} \textcolor{blue}{Description]}. The periodic lattice of this MOF is \textcolor{blue}{[Lattice} \textcolor{blue}{Parameters]}. 

Local properties of each block, whose principal axis is aligned with [1,0,0] in the local frame, are listed below.

[0] SMILES=\textcolor{blue}{[SMILES]}, molecular weight=\textcolor{blue}{[Molecular} \textcolor{blue}{Weight]}, PCA span=\textcolor{blue}{[PCA Span]} 

[1] ...

Global placement of each building block is described as below.

[0] Translation=\textcolor{blue}{[Translation]}, rotation=\textcolor{blue}{[Rotation]}, rotated main axis=\textcolor{blue}{[Rotated Principal Axis]}

[1] ...
\end{tcolorbox}

\noindent The complete prompt templates and a detailed description of the dataset construction process are provided in the Appendix \ref{app:prompt}. Based on this constructed dataset, we train a pre-trained LLM with the standard next-token prediction objective.

\subsection{Structural Supervised Fine-tuning }\label{method_3}
We perform structural SFT to equip the LLM with the ability to autoregressively assemble MOF structures.
In this stage, the LLM takes as input an instruction $I$ which includes the task description, the output format, and a set of building blocks, and is fine-tuned to generate a response $R$ consisting of lattice parameters and block-wise roto-translations:
\begin{tcolorbox}[
colback=white, 
colframe=black!20, 
fontupper=\ttfamily\small, 
halign=justify,
left=1.5mm,
right=0mm
] 
\textbf{Instruction:} \textcolor{blue}{[Task Description]}. Given the SMILES of all MOF building blocks, predict the complete 3D structure configuration including the lattice parameters in the format 'a b c $\alpha$ $\beta$ $\gamma$', and for each building block (maintaining the exact same order as provided in the input):

\ - Translation vector ([tx ty tz]): The position of the building block's center within the unit cell, expressed in fractional coordinates.

\ - Rotation angles ([roll pitch yaw]): The orientation of the building block, represented in radians using Euler angles.

Output format: \textcolor{blue}{[Output Format]}

\textcolor{blue}{[SMILES]} \textcolor{blue}{[SMILES]} ...

\textbf{Response:} \textcolor{red}{[Lattice Parameters]}

[0] \textcolor{red}{[Translation] [Rotation]} 

[1] \textcolor{red}{[Translation] [Rotation]}

[2] ...
\end{tcolorbox}
\noindent The model is fine-tuned by minimizing the following generation loss:
\begin{equation}
    \mathcal{L}(\theta)=-\sum_{t=1}^{|R|}\mathrm{log}\ \pi_\theta(R_{t}|I,R_{< t}),
\end{equation}
where $\pi_\theta$ denotes the LLM parameterized by $\theta$, initialized from the spatial-aware CPT model.

\subsection{Matching-driven Reinforcement Learning}\label{method_4}
While SFT enables effective block-wise placement by token-level cross-entropy loss, it inherently fails to enforce global structural plausibility.
Consequently, the generated MOFs may exhibit structural instabilities.
To address this, we employ reinforcement learning (RL) with global structural reward to further refine the SFT model.
Specifically, we adopt the Soft Adaptive Policy Optimization (SAPO) framework, a soft and adaptive generalization of Group Relative Policy Optimization (GRPO)~\citep{guo2025deepseek,shao2024deepseekmath}. This group-based RL paradigm derives informative learning signals from relative comparisons among multiple outputs, even when most are suboptimal.

For each instruction $I$, the model first samples a group of $G$ candidate responses $\{R^{(1)}, \dots, R^{(G)}\}$, which are subsequently parsed into the corresponding MOF structures $\{\mathcal{S}^{(1)}, \dots, \mathcal{S}^{(G)}\}$.
We then evaluate the quality of each structure $\mathcal{S}^{(i)}$ by comparing it with the ground-truth stable reference $\mathcal{S}_{gt}$. 
To quantify this alignment, we compute a reward based on structural matching fidelity. Let $\tau$ denote the minimum site tolerance at which the two structures are considered matched, as determined using the \texttt{StructureMatcher} class in \texttt{pymatgen}~\citep{ong2013python}. The reward $r^{(i)}$ function is defined as:
% \begin{equation}
% r^{(i)} = \left\{
% \begin{array}{l @{\quad} l} 
% -1, & \quad \text{if } \mathcal{S}^{(i)} \text{ cannot be parsed,} \\[1ex]
% \multicolumn{2}{l}{1 + \frac{1}{2} \exp\left( -4 \cdot \mathtt{RMSE}(\mathcal{S}^{(i)}, \mathcal{S}_{gt}) \right)}, \\[1ex]
%    & \quad \text{if } \tau\leq 0.5, \\[1ex]
% 0.6, & \quad \text{if } 0.5<\tau\leq 0.75, \\[1ex]
% 0.3, & \quad \text{if } 0.75<\tau\leq 1.0, \\[1ex]
% 0, & \quad \text{otherwise.}
% \end{array}
% \right.
% \end{equation}
\begin{equation}
r^{(i)} = \left\{
\begin{aligned}
& -1, \qquad\qquad \text{if } \mathcal{S}^{(i)}\ \text{parsing failed,} \\[1ex]
& 1 + 0.5e^{-4 \cdot \mathtt{RMSE}(\mathcal{S}^{(i)}, \mathcal{S}_{gt})}, \quad \text{if } \tau\leq 0.5, \\[1ex]
& 0.6,\ \qquad\qquad \text{if } 0.5<\tau\leq 0.75, \\[1ex]  
& 0.3,\ \qquad\qquad \text{if } 0.75<\tau\leq 1.0, \\[1ex]
& 0, \qquad\qquad\quad \text{otherwise.}
\end{aligned}
\right.
\label{eq:reward}
\end{equation}
\noindent where $\mathtt{RMSE}(\cdot, \cdot)$ represents the root mean square distance between matched sites of the two structures. 
We provide more details about the reward calculation in Appendix \ref{app:rl}.
This reward function encourages high-precision generation ($\tau \leq 0.5$) by providing an additional bonus for low geometric error, while assigning no bonus to suboptimal structures ($0.5 < \tau \leq 1.0$) to promote training stability.

\begin{table*}[t]
\centering
\caption{
Structure prediction accuracy. 
$\mathrm{stol}$ represents the site tolerance for matching criteria. 
MR is the match rate and RMSE is the root mean square error between generated structures and the ground-truths.
The reported time is the average per structure.
}
\label{tb:main_res}

\renewcommand{\arraystretch}{0.85}
\resizebox{0.78\textwidth}{!}{
\begin{tabular}{l c cc cc c}
\toprule
 & \multirow{2}{*}{\# of samples} 
 & \multicolumn{2}{c}{$\mathrm{stol} = 0.5$} 
 & \multicolumn{2}{c}{$\mathrm{stol} = 1.0$}
 & \multirow{2}{*}{Avg. time (s)$\downarrow$} 
 \\
\cmidrule(lr){3-4}\cmidrule(lr){5-6}
 & 
 & MR (\%)$\uparrow$ & RMSE$\downarrow$ 
 & MR (\%)$\uparrow$ & RMSE$\downarrow$
 &
 \\
\midrule
\rowcolor{gray!30}
\multicolumn{7}{c}{\textbf{Denoising-based}} \\
\midrule
\multirow{2}{*}{DiffCSP~\citep{jiao2023crystal}} 
 & 1 & 0.09 & 0.3961 & 23.12 & 0.8294 & 5.37 \\
 & 5 & 0.34 & 0.3848 & 38.94 & 0.7937 & 26.85 \\
\midrule
\multirow{2}{*}{MOFFlow~\citep{kimmofflow}} 
& 1 & 31.69 & 0.2820 & 87.46 & 0.5183 & 0.21 \\
& 5 & 44.75 & 0.2694 & \textbf{100.0} & 0.4645 & 1.07 \\
\midrule
 \multirow{2}{*}{MOF-BFN~\citep{jiao2025mof}} 
 & 1 & 35.27 & 0.2735 & 92.99 & 0.5000 & \textbf{0.04} \\
 & 5 & 53.51 & 0.2498 & 98.37 & 0.4117 & 0.21 \\
\midrule
\rowcolor{gray!30}
\multicolumn{7}{c}{\textbf{LLM-based}} \\
\midrule
\multirow{2}{*}{PLaID++~\citep{xu2025plaid++}}
 & 1 & 0.09 & 0.3943 & 6.22 & 0.7794 & 0.41 \\
 & 5 & 0.16 & 0.4067 & 19.94 & 0.7981 & 2.07 \\
\midrule
\multirow{2}{*}{\methodname (Ours)} 
 & 1 & 35.78	& 0.2612 & 93.25 & 0.4984 & \textbf{0.04} \\
 & 5 & \textbf{53.64} & \textbf{0.2383} & 98.48 & \textbf{0.4080} & 0.18 \\
\bottomrule
\end{tabular}
}
\end{table*}

The model is then updated by maximizing the SAPO objective:
\begin{equation}
\mathcal{J}_{\mathrm{SAPO}}(\theta)=\frac{1}{G} \sum_{i=1}^{G} \frac{1}{\left|R^{(i)}\right|} \sum_{t=1}^{\left|R^{(i)}\right|} f_{i, t}\left(\bar{r}_{i, t}(\theta)\right) A^{(i)}.
\end{equation}
Here, $\bar{r}_{i,t}(\theta)$ is the importance ratio between the current policy $\pi_\theta$ and a reference policy $\pi_{\theta_{\mathrm{ref}}}$, from which the responses are sampled, at time step $t$:
\begin{equation}
\bar{r}_{i,t}(\theta)=
\frac{\pi_\theta(R_t^{(i)}\mid I, R_{<t}^{(i)})}
{\pi_{\theta_{\mathrm{ref}}}(R_t^{(i)}\mid I, R_{<t}^{(i)})},
\end{equation}
and $A^{(i)}$ is a group-normalized advantage:
\begin{equation}
A^{(i)}=\frac{r^{(i)} - \mathrm{mean}\big([r^{(i)}]_{i=1}^{G}\big)}{\mathrm{std}\big([r^{(i)}]_{i=1}^{G}\big)}.
\end{equation}
The function $f_{i,t}(\cdot)$ is a smooth, temperature-controlled gating function that replaces the hard clipping used in GRPO, enabling stable and informative policy updates by preserving useful learning signals from moderately off-policy tokens:
\begin{equation}
\begin{aligned}
f_{i,t}(x)=
\sigma\big(\tau_{i,t}(x-1)\big)\cdot\frac{4}{\tau_{i,t}},\quad
\tau_{i,t}=
\begin{cases}
\tau_{\mathrm{pos}}, & \text{if } A^{(i)} > 0, \\
\tau_{\mathrm{neg}}, & \text{otherwise}.
\end{cases}
\end{aligned}
\end{equation}
where $\tau_{\mathrm{pos}}$ and $\tau_{\mathrm{neg}}$ are temperate paramters for postive and negative tokens, and $\sigma(x)=1/(1+e^{-x})$ is the sigmoid function.

\section{Experiments}
\begin{figure*}[t]
  \includegraphics[width=.85\textwidth]{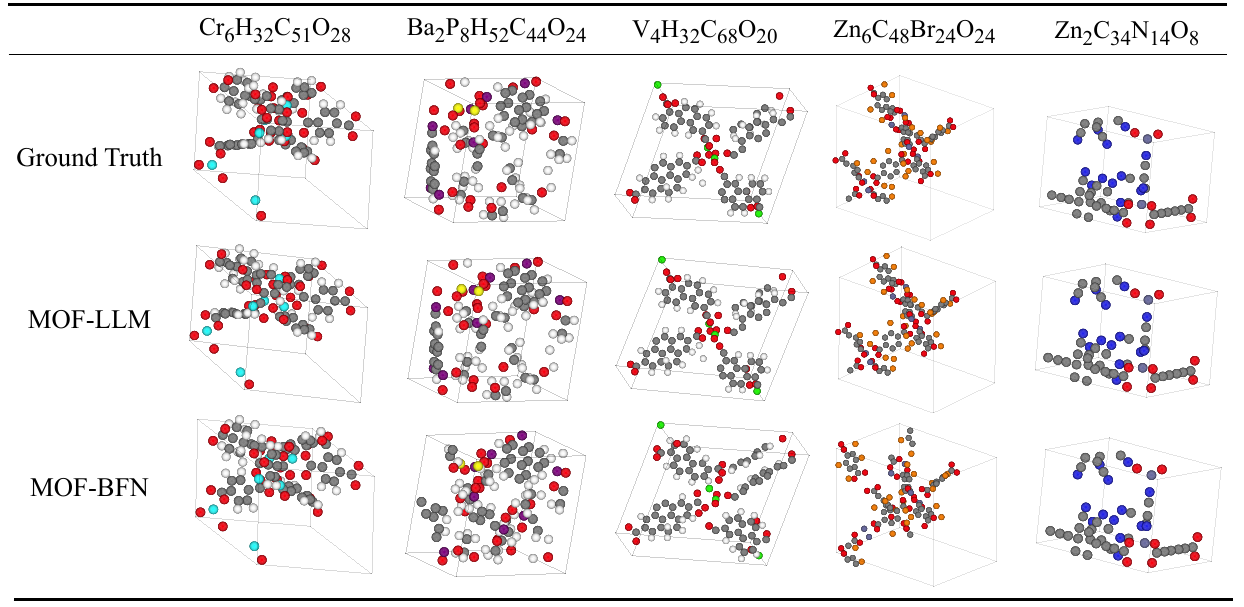}
  \caption{Visualization of five selected samples where \methodname's predictions match the ground truth. The unit cells are rescaled for visual clarity. \methodname accurately captures the atomic positions and lattice parameters.}
  \label{fig:gen_sample}
\end{figure*}
In this section, we evaluate \methodname from multiple perspectives. First, we report structure prediction accuracy in \Cref{exp_1}, demonstrating the superior performance of \methodname compared with both existing denoising-based and LLM-based methods. 
Next, in \Cref{exp_2}, we assess the structural properties of the predicted outputs, showing that \methodname successfully captures key characteristics of MOF structures. 
Finally, we analyze the contribution and effectiveness of each stage within \methodname.

\subsection{Structure Prediction}\label{exp_1}
\textbf{Dataset.} We use the dataset introduced by \citet{boyd2019data}, which contains 324,426 computationally generated hypothetical MOF structures constructed through a topology-based assembly of building blocks. 
The dataset was originally developed for large-scale screening of MOFs for CO$_2$ capture under realistic flue gas conditions.
In practice, we adopt the preprocessed version of this dataset\footnote{\url{https://zenodo.org/records/15187230}} provided by \citet{kimmofflow}, where all MOF structures are converted to block-level representation detailed in \Cref{sec:mof_repr} using the \texttt{metal-oxo} algorithm~\citep{bucior2019identification}. Structures containing more than 200 blocks are discarded, and the remaining data are split into training, validation, and test sets with an 8:1:1 ratio.

\textbf{Baselines.} We compare \methodname against two types of baselines: denoising-based methods and LLM-based methods.
For the former, we include DiffCSP~\citep{jiao2023crystal}, a full-atom diffusion model, as well as MOFFLOW~\citep{kimmofflow} and MOF-BFN~\citep{jiao2025mof}, both of which are block-level generative models specifically designed for MOFs.
For LLM-based methods, we benchmark against PLaID++~\citep{xu2025plaid++}, a full-atom crystal generation model based on Wyckoff text presentation. Since PLaID++ is trained on bulk materials and based on Qwen-2.5, we reimplement it with Qwen-3 on our dataset for a fair comparison. Details are provided in Appendix \ref{app:baseline}.

\textbf{Metrics.} We use the match rate (MR) and root mean square error (RMSE) for evaluation.
A predicted structure matches the ground truth if its deviations in site positions, lattice lengths and angles all lie within predefined tolerance thresholds.
We use the \texttt{StructureMatcher} class in \texttt{pymatgen} and employ two sets of tolerances parameters (\texttt{stol}, \texttt{ltol}, \texttt{atol}): $(0.5,0.3,10.0)$ and $(1.0,0.3,10.0)$, following ~\citet{kimmofflow,jiao2025mof}.
For each test case, we generate 1 and 5 structures, and the MR is defined as the proportion that at least one candidate matches the ground truth, while RMSE is computed as the per-atom root mean squared displacement between the ground truth and the best-matching prediction.

\textbf{Results.} \Cref{tb:main_res} presents the main results.
As expected, all-atom generation methods (DiffCSP and PLaID++) yield poor performance, achieving near-zero MR under the strict criterion ($\mathrm{stol}=0.5$) and remaining suboptimal even with a more lenient condition of $\mathrm{stol}=1.0$.
This highlights the inherent limitations of all-atom approaches for predicting complex MOF structures.
Notably, PLaID++ underperforms DiffCSP, likely due to the deficiency of LLMs to process extensive floating-point numerical data in atom-level representations, further motivating block-level modeling.
In contrast, \methodname outperforms current block-level baselines (MOFFlow and MOF-BFN), consistently achieving higher MR and lower RMSE across all settings, demonstrating superior structural accuracy. 
\Cref{fig:gen_sample} visualizes some predicted MOF structures of \methodname.

We further evaluate sampling efficiency. Unlike denoising-based methods that require iterative sampling, \methodname generates samples autoregressively in a single pass, reducing inference time to 0.04 seconds per structure.
Notably, \methodname maintains a lower total runtime in the five-sample evaluation compared to MOF-BFN (0.18 s vs. 0.21 s), indicating superior efficiency as sampling demand increases.
Coupled with ongoing advances in LLM acceleration~\citep{kwon2023efficient,zheng2024sglang}, this efficiency underscores the strong potential of our approach for practical high-throughput MOF discovery.

\subsection{Geometric Property Evaluation}\label{exp_2}
In this section, we evaluate the geometric properties of the predicted MOF structures, which are key determinants of practical performance. For example, the pore limiting diameter (PLD), which characterizes the narrowest constriction along a pore channel, governs molecular adsorption selectivity and directly affects performance in applications such as carbon capture. 
We utilized the \texttt{Zeo++}~\citep{willems2012algorithms} to compute common geometric descriptors, including volumetric surface area (VSA), gravimetric surface area (GSA), accessible volume (AV), unit cell volume (UCV), void fraction (VF), PLD, largest cavity diameter (LCD), and density (DST). 

\begin{table}[htbp]
\centering
\caption{Geometric property evaluation. RMSE computed between the predicted and ground-truth structures, with baseline results from \citet{jiao2025mof} and \citet{kimmofflow}.}
\label{tab:geo_prop}

\resizebox{0.93\linewidth}{!}{
\begin{tabular}{lccc}
\toprule
 & \multicolumn{3}{c}{RMSE$\downarrow$} \\
\cmidrule(lr){2-4}
 & \methodname & MOF-BFN & MOFFlow \\
\midrule
VSA (m$^2$/cm$^3$) & 251.3 & \textbf{232.8} & 264.5 \\
GSA (m$^2$/g) & 266.2 & \textbf{247.5} & 331.6 \\
AV (\AA$^3$) & \textbf{198.8} & 315.4 & 530.5 \\
UCV (\AA$^3$) & \textbf{168.8} & 312.0 & 569.5 \\
VF & 0.0212 & \textbf{0.0187} & 0.0285 \\
PLD (\AA) & 0.9303 & \textbf{0.9072} & 1.0616 \\
LCD (\AA) & 0.9509 & \textbf{0.9257} & 1.1083 \\
DST (g/cm$^3$) & \textbf{0.0137} & 0.0185 & 0.0442 \\
\bottomrule
\end{tabular}
}
\end{table}

\Cref{tab:geo_prop} presents the RMSE between the geometric properties of the predicted structures and the ground-truth references on the test set. Notably, \methodname performs competitively with MOF-BFN, the state-of-the-art denoising-based generative models, while significantly outperforming MOFFlow. 
Overall, these results indicate that our framework effectively captures essential spatial characteristics and preserves structural integrity, highlighting its promise for practical functional MOF discovery.

\subsection{Scalability Evaluation}
Traditional CSP methods typically scale poorly to realistically sized systems~\cite{neumann2024orbfastscalableneural,rhodes2025orbv3atomisticsimulationscale,jiao2023crystal}. In this section, we evaluate the scalability of MOF-LLM as the MOF system size grows. Specifically, we assess the MR with an increasing number of atoms and building blocks in the unit cell. MOF-LLM is compared with MOF-BFN, the best performing baseline, under tolerance parameters (0.5, 0.3, 10.0).

\begin{figure}[htbp]
  \includegraphics[width=\linewidth]{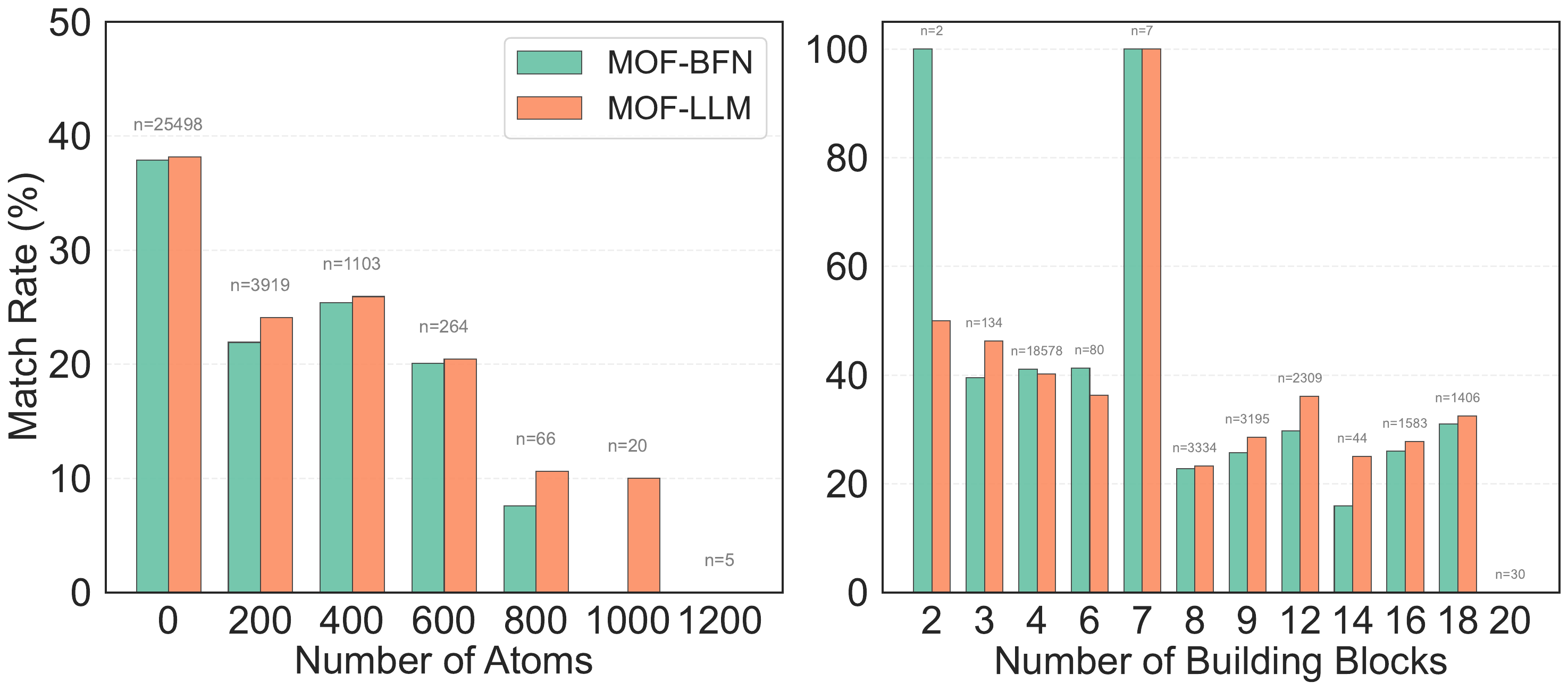}
  \caption{Scalability evaluation. MOF structures are binned according to the number of atoms (\textit{left}) and building blocks (\textit{right}). For each bin, the match rate is computed for both MOF-LLM and MOF-BFN. Samples exceeding 1400 atoms or 20 building blocks are excluded, where both methods yielded a zero MR.}
  \label{fig:scaling}
\end{figure}

\Cref{fig:scaling} shows the scalability of both models. For the atom-based analysis, samples are partitioned into bins with intervals of $(t, t+200]$. MOF-LLM consistently achieves higher MR than MOF-BFN across all bins. Notably, the performance gap widens for larger systems, particularly those containing more than 800 atoms. A similar trend is observed when binned by the number of building blocks, with MOF-LLM showing a clear advantage for structures comprising more than seven blocks. These results demonstrate that MOF-LLM scales favorably with system size, highlighting its potential for predicting complex MOF structures.

\subsection{Ablation Study}
\textbf{Euler angles provide an LLM-friendly rotation representation.} Beyond Euler angles, rotations can be encoded using other compact representations such as unit quaternions and axis–angle vectors.
We specifically compared Euler angles with the axis-angle vector, which is defined as the product of a rotation angle and a unit axis (see Appendix \ref{app:rotation}) and has been successfully employed in autoregressive molecular structure generation~\citep{fufragment}. 
As shown in \Cref{tb:ab_rotation}, the Euler angle-based representation yields a substantial structure prediction accuracy over the axis-angle representation. 
We speculate that Euler angles provide a more intuitive format for LLMs, whereas axis-angle representations demand stronger spatial priors, which may hinder the LLM’s precise prediction of block orientations.
\begin{table}[htbp]
\centering
\caption{
\textbf{Ablation study on rotation representations}. 
The axis–angle rotation vector is from \citet{fufragment}. 
Both models are fine-tuned via SFT from a Qwen-3 8B base model.
}
\label{tb:ab_rotation}

\resizebox{0.98\linewidth}{!}{
\begin{tabular}{l cc cc}
\toprule
 & \multicolumn{2}{c}{$\mathrm{stol} = 0.5$} 
 & \multicolumn{2}{c}{$\mathrm{stol} = 1.0$}
 \\
\cmidrule(lr){2-3}\cmidrule(lr){4-5} 
 & MR (\%)$\uparrow$ & RMSE$\downarrow$ 
 & MR (\%)$\uparrow$ & RMSE$\downarrow$
 \\
\midrule
Euler angle & 34.60 & 0.2644 & 92.33 & 0.5058 \\
Axis-angle vector & 31.82 & 0.2727 & 91.11 & 0.5190 \\
\bottomrule
\end{tabular}
}
\end{table}

\begin{figure}[t]
  \includegraphics[width=0.95\linewidth]{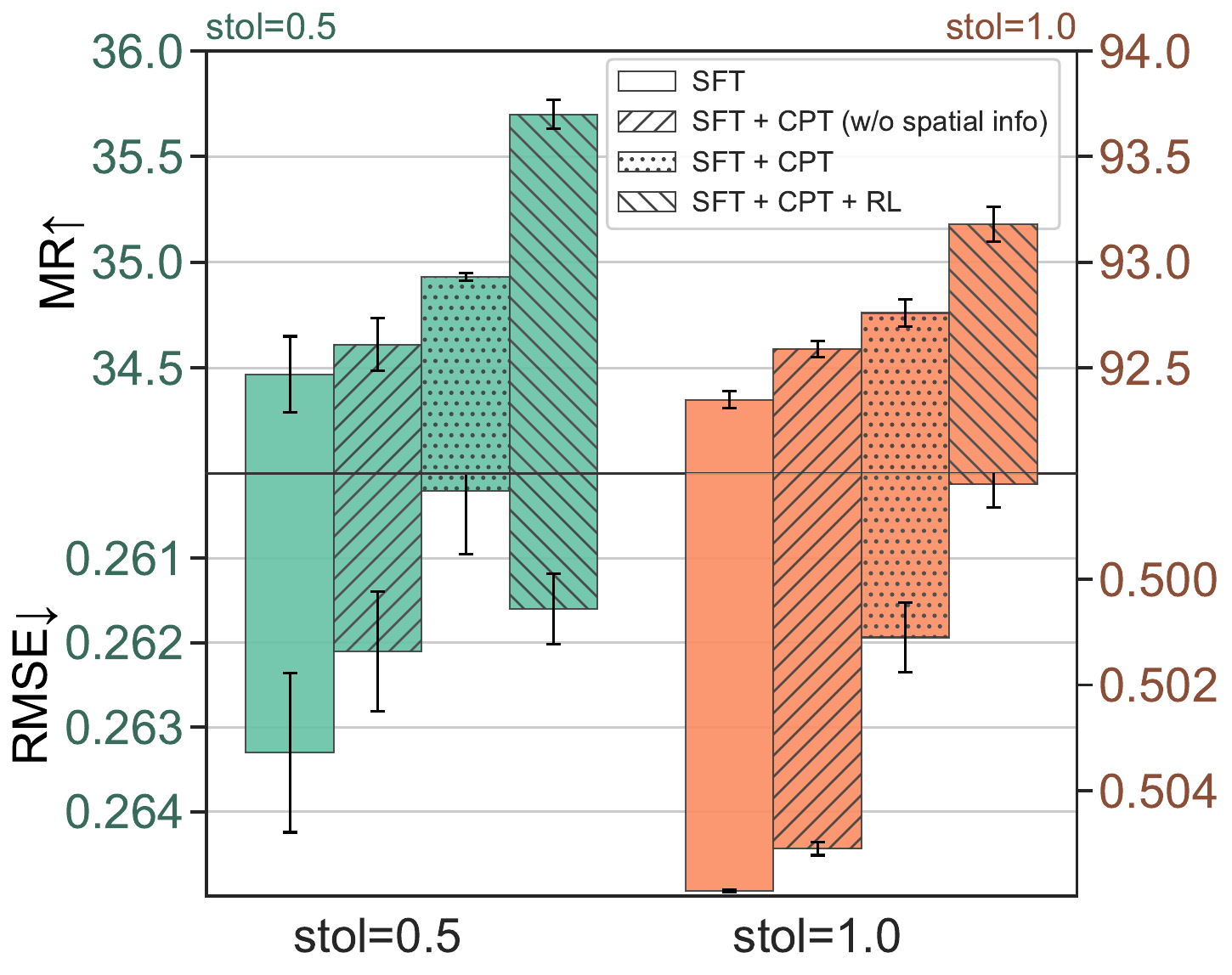}
  \caption{Ablation study of key design components. CPT (w/o spatial info) denotes CPT using prompts without topology codes, molecular weight, PCA spans, and rotated axes. We report the standard deviation across a total of 9 runs (3 random seeds, each sampled 3 times).}
  \label{fig:abaltion}
\end{figure}

\begin{figure}[t]
  \includegraphics[width=0.96\linewidth]{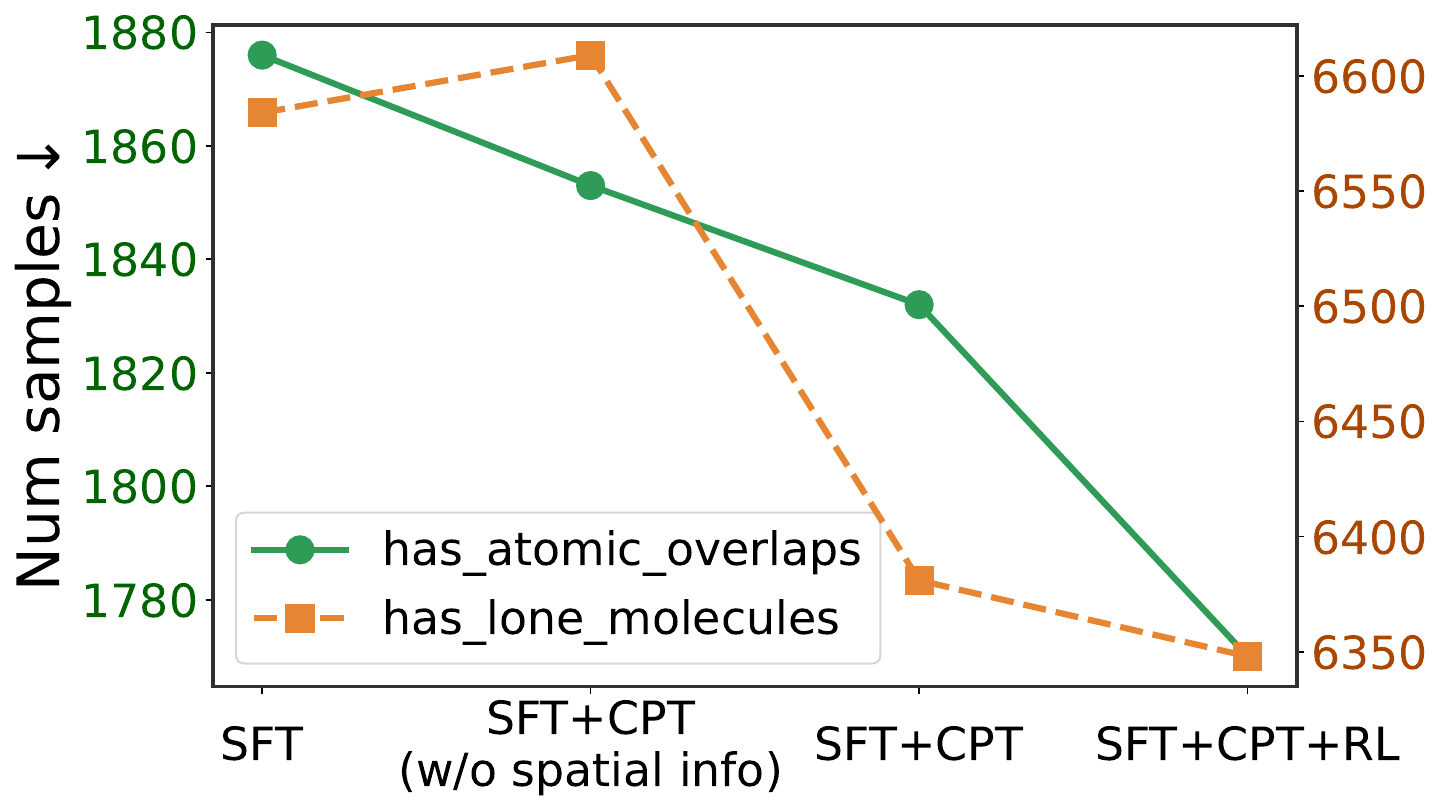}
  \caption{Structural validity evaluation. The number of generated structures flagged with has\_atomic\_overlaps and has\_lone\_molecules is reported, corresponding to block collisions and disconnections, respectively.}
  \label{fig:mofchecker}
\end{figure}

\textbf{Spatial-aware CPT greatly enhances LLMs' spatial reasoning capabilities.} As shown in \Cref{fig:abaltion}, incorporating our proposed CPT stage yields a significant MR improvement over the pure SFT model. 
To isolate the contribution of explicit spatial knowledge, we conducted an ablation study in which all spatial and geometric descriptors, including the topology code, molecular weight, PCA spans, and rotated axis, were removed from the CPT prompts (`w/o spatial info'). This results in a clear decline in MR and a notable increase in RMSE.
To better characterize the spatial failure modes, we further evaluate the generated structures using two \texttt{MOFChecker}~\citep{jin2025mofchecker} validity metrics in \Cref{fig:mofchecker}: \texttt{has\_atomic\_overlaps} and \texttt{has\_lone\_molecules}, which indicate steric clashes and topological errors, respectively. Spatial-aware CPT substantially reduces both types of structural invalidity compared with SFT and CPT without spatial priors.
These findings demonstrate the effectiveness of the spatial-aware strategy in enabling LLMs to interpret complex 3D geometries.

\begin{figure}[t]
  \includegraphics[width=\linewidth]{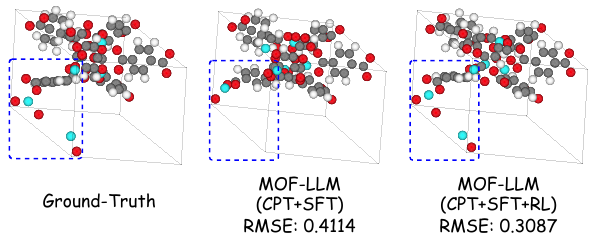}
  \caption{Visualization of the generated structures of Cr$_6$H$_{32}$C$_{51}$O$_{28}$. The RL stage substantially improves the prediction accuracy. The dashed boxes highlight regions with the greatest prediction improvements.}
  \label{fig:sft_rl_comp}
\end{figure}

\textbf{Matching-driven RL constitutes a critical component.} As shown in \Cref{fig:abaltion}, incorporating the RL stage achieves the best overall performance, yielding a pronounced improvement in predictive accuracy. 
Notably, the RL stage significantly reduces the number of structurally implausible structures, effectively mitigating errors such as block collisions and disconnections (\Cref{fig:mofchecker}).
A subsequent qualitative analysis in \Cref{fig:sft_rl_comp} further corroborates that this RL refinement aligns the generated structures more closely with the ground truth.
Overall, these results demonstrate that SAPO effectively steers the generation policy toward high-reward distribution, thereby substantially improving the accuracy of the predicted structures.

\section{Conclusion and Discussion}
In this work, we introduce \methodname, the first framework that successfully adapts LLMs for predicting the structure of MOFs. By employing a three-stage training pipeline integrating spatial-aware CPT, structural SFT, and matching-driven RL, our approach substantially enhances the spatial reasoning capabilities of LLMs for complex building-block assembly. Comprehensive evaluations demonstrate the effectiveness and efficiency of our framework, while ablation studies confirm the critical roles of each component in enabling robust geometric understanding and structural stability. 
% Together, these results highlight the potential of \methodname for MOF discovery.

From a broader perspective, this work contributes to reticular chemistry and materials science by introducing an efficient paradigm for accelerated MOF discovery, effectively overcoming the scalability bottlenecks of traditional workflows. 
Specifically, \methodname tackles the challenge of structural complexity of MOFs that renders conventional first-principles methods computationally prohibitive, offering an accurate and highly efficient alternative through LLM-based modeling. 
Applying LLMs to MOF structure prediction is inherently challenging due to their limited spatial reasoning at microscopic scales. Our meticulously designed post-training framework overcomes this and establishes a principled pathway for adapting generic LLMs to intricate scientific systems.
We anticipate that \methodname can serve as a foundational stepping stone toward next-generation MOF discovery, where natural language prompting directly enables versatile MOF design.

\section{Limitations and Ethical Considerations}
\textbf{Limitations.} While \methodname enables accurate and rapid prediction of MOF structures, it still has several limitations. First, following established protocols~\citep{kimmofflow,jiao2025mof}, we treat building blocks as rigid bodies and assume the availability of ground-truth block structures. Extending the framework to model conformational flexibility could further enhance practicality. Second, as discussed in \Cref{method_1}, flattening 3D blocks into 1D SMILES strings inherently simplifies geometric details. Future research could delve deeper into advanced tokenization methods that natively preserve 3D geometries. Finally, while our model directly generates solutions, integrating explicit Chain-of-Thought reasoning offers a promising direction for improving interpretability and complex spatial planning.

\textbf{Ethical considerations.} \methodname is designed to support beneficial applications like carbon capture. However, its ability to rapidly predict complex MOF structures could be misused for the design of materials intended to store or transport hazardous substances. 
Addressing these risks requires responsible governance, careful oversight, and strict adherence to established ethical guidelines by the research community to ensure safe use of the technology.
In addition, ongoing advances in LLMs' safety alignment~\citep{shen2023large} techniques may offer a complementary avenue to help mitigate potential misuse and encourage responsible deployment.

% \begin{acks}
% \end{acks}

\section*{GenAI Disclosure}
We used generative AI tools (ChatGPT and Gemini; accessed in 2026) for language polishing and rephrasing throughout the manuscript. These tools were not used to generate new technical content, experimental results, figures, data, code, or references. All content was reviewed, verified, and finalized by the authors, who take full responsibility for the paper.

\section*{Acknowledgments}
This work was supported by the Natural Science Foundation of China (Grant No. 62406170). We also acknowledge the support from Chemistry and Biomedicine Innovation Center (ChemBIC) and the AI \& AI for Science Project of Nanjing University.

\bibliographystyle{ACM-Reference-Format}
\bibliography{sample-base}


\begin{thebibliography}{52}

%%% the Web address.

\ifx \showCODEN    \undefined \def \showCODEN     #1{\unskip}     \fi
\ifx \showISBNx    \undefined \def \showISBNx     #1{\unskip}     \fi
\ifx \showISBNxiii \undefined \def \showISBNxiii  #1{\unskip}     \fi
\ifx \showISSN     \undefined \def \showISSN      #1{\unskip}     \fi
\ifx \showLCCN     \undefined \def \showLCCN      #1{\unskip}     \fi
\ifx \shownote     \undefined \def \shownote      #1{#1}          \fi
\ifx \showarticletitle \undefined \def \showarticletitle #1{#1}   \fi
\ifx \showURL      \undefined \def \showURL       {\relax}        \fi
% invisible to TeX
\providecommand\bibfield[2]{#2}
\providecommand\bibinfo[2]{#2}
\providecommand\natexlab[1]{#1}
\providecommand\showeprint[2][]{arXiv:#2}

\bibitem[Antunes et~al\mbox{.}(2024)]%
        {antunes2024crystal}
\bibfield{author}{\bibinfo{person}{Luis~M Antunes}, \bibinfo{person}{Keith~T Butler}, {and} \bibinfo{person}{Ricardo Grau-Crespo}.} \bibinfo{year}{2024}\natexlab{}.
\newblock \showarticletitle{Crystal structure generation with autoregressive large language modeling}.
\newblock \bibinfo{journal}{\emph{Nature Communications}} \bibinfo{volume}{15}, \bibinfo{number}{1} (\bibinfo{year}{2024}), \bibinfo{pages}{10570}.
\newblock


\bibitem[Badrinarayanan et~al\mbox{.}(2025)]%
        {badrinarayanan2025mofgpt}
\bibfield{author}{\bibinfo{person}{Srivathsan Badrinarayanan}, \bibinfo{person}{Rishikesh Magar}, \bibinfo{person}{Akshay Antony}, \bibinfo{person}{Radheesh~Sharma Meda}, {and} \bibinfo{person}{Amir Barati~Farimani}.} \bibinfo{year}{2025}\natexlab{}.
\newblock \showarticletitle{MOFGPT: Generative design of metal--organic frameworks using language models}.
\newblock \bibinfo{journal}{\emph{Journal of Chemical Information and Modeling}} \bibinfo{volume}{65}, \bibinfo{number}{17} (\bibinfo{year}{2025}), \bibinfo{pages}{9049--9060}.
\newblock


\bibitem[Boyd et~al\mbox{.}(2019)]%
        {boyd2019data}
\bibfield{author}{\bibinfo{person}{Peter~G Boyd}, \bibinfo{person}{Arunraj Chidambaram}, \bibinfo{person}{Enrique Garc{\'\i}a-D{\'\i}ez}, \bibinfo{person}{Christopher~P Ireland}, \bibinfo{person}{Thomas~D Daff}, \bibinfo{person}{Richard Bounds}, \bibinfo{person}{Andrzej G{\l}adysiak}, \bibinfo{person}{Pascal Schouwink}, \bibinfo{person}{Seyed~Mohamad Moosavi}, \bibinfo{person}{M~Mercedes Maroto-Valer}, {et~al\mbox{.}}} \bibinfo{year}{2019}\natexlab{}.
\newblock \showarticletitle{Data-driven design of metal--organic frameworks for wet flue gas CO2 capture}.
\newblock \bibinfo{journal}{\emph{Nature}} \bibinfo{volume}{576}, \bibinfo{number}{7786} (\bibinfo{year}{2019}), \bibinfo{pages}{253--256}.
\newblock


\bibitem[Bucior et~al\mbox{.}(2019)]%
        {bucior2019identification}
\bibfield{author}{\bibinfo{person}{Benjamin~J Bucior}, \bibinfo{person}{Andrew~S Rosen}, \bibinfo{person}{Maciej Haranczyk}, \bibinfo{person}{Zhenpeng Yao}, \bibinfo{person}{Michael~E Ziebel}, \bibinfo{person}{Omar~K Farha}, \bibinfo{person}{Joseph~T Hupp}, \bibinfo{person}{J~Ilja Siepmann}, \bibinfo{person}{Al{\'a}n Aspuru-Guzik}, {and} \bibinfo{person}{Randall~Q Snurr}.} \bibinfo{year}{2019}\natexlab{}.
\newblock \showarticletitle{Identification schemes for metal--organic frameworks to enable rapid search and cheminformatics analysis}.
\newblock \bibinfo{journal}{\emph{Crystal Growth \& Design}} \bibinfo{volume}{19}, \bibinfo{number}{11} (\bibinfo{year}{2019}), \bibinfo{pages}{6682--6697}.
\newblock


\bibitem[Castro~Nascimento and Pimentel(2023)]%
        {castro2023large}
\bibfield{author}{\bibinfo{person}{Cayque~Monteiro Castro~Nascimento} {and} \bibinfo{person}{Andr{\'e}~Silva Pimentel}.} \bibinfo{year}{2023}\natexlab{}.
\newblock \showarticletitle{Do large language models understand chemistry? a conversation with chatgpt}.
\newblock \bibinfo{journal}{\emph{Journal of Chemical Information and Modeling}} \bibinfo{volume}{63}, \bibinfo{number}{6} (\bibinfo{year}{2023}), \bibinfo{pages}{1649--1655}.
\newblock


\bibitem[Dao(2023)]%
        {dao2023flashattention}
\bibfield{author}{\bibinfo{person}{Tri Dao}.} \bibinfo{year}{2023}\natexlab{}.
\newblock \showarticletitle{Flashattention-2: Faster attention with better parallelism and work partitioning}.
\newblock \bibinfo{journal}{\emph{arXiv preprint arXiv:2307.08691}} (\bibinfo{year}{2023}).
\newblock


\bibitem[DeepSeek-AI(2024)]%
        {deepseekai2024deepseekv3technicalreport}
\bibfield{author}{\bibinfo{person}{DeepSeek-AI}.} \bibinfo{year}{2024}\natexlab{}.
\newblock \bibinfo{title}{DeepSeek-V3 Technical Report}.
\newblock
\showeprint[arxiv]{2412.19437}~[cs.CL]
\urldef\tempurl%
\url{https://arxiv.org/abs/2412.19437}
\showURL{%
\tempurl}


\bibitem[Ding et~al\mbox{.}(2025)]%
        {ding2025matexpert}
\bibfield{author}{\bibinfo{person}{Qianggang Ding}, \bibinfo{person}{Santiago Miret}, {and} \bibinfo{person}{Bang Liu}.} \bibinfo{year}{2025}\natexlab{}.
\newblock \showarticletitle{MatExpert: Decomposing Materials Discovery By Mimicking Human Experts}. In \bibinfo{booktitle}{\emph{The Thirteenth International Conference on Learning Representations}}.
\newblock


\bibitem[Dubey et~al\mbox{.}(2024)]%
        {dubey2024llama}
\bibfield{author}{\bibinfo{person}{Abhimanyu Dubey}, \bibinfo{person}{Abhinav Jauhri}, \bibinfo{person}{Abhinav Pandey}, \bibinfo{person}{Abhishek Kadian}, \bibinfo{person}{Ahmad Al-Dahle}, \bibinfo{person}{Aiesha Letman}, \bibinfo{person}{Akhil Mathur}, \bibinfo{person}{Alan Schelten}, \bibinfo{person}{Amy Yang}, \bibinfo{person}{Angela Fan}, {et~al\mbox{.}}} \bibinfo{year}{2024}\natexlab{}.
\newblock \showarticletitle{The llama 3 herd of models}.
\newblock \bibinfo{journal}{\emph{arXiv e-prints}} (\bibinfo{year}{2024}), \bibinfo{pages}{arXiv--2407}.
\newblock


\bibitem[Fu et~al\mbox{.}(2024)]%
        {fufragment}
\bibfield{author}{\bibinfo{person}{Cong Fu}, \bibinfo{person}{Xiner Li}, \bibinfo{person}{Blake Olson}, \bibinfo{person}{Heng Ji}, {and} \bibinfo{person}{Shuiwang Ji}.} \bibinfo{year}{2024}\natexlab{}.
\newblock \showarticletitle{Fragment and Geometry Aware Tokenization of Molecules for Structure-Based Drug Design Using Language Models}. In \bibinfo{booktitle}{\emph{The Thirteenth International Conference on Learning Representations}}.
\newblock


\bibitem[Fu et~al\mbox{.}(2023)]%
        {fumofdiff}
\bibfield{author}{\bibinfo{person}{Xiang Fu}, \bibinfo{person}{Tian Xie}, \bibinfo{person}{Andrew~Scott Rosen}, \bibinfo{person}{Tommi~S Jaakkola}, {and} \bibinfo{person}{Jake~Allen Smith}.} \bibinfo{year}{2023}\natexlab{}.
\newblock \showarticletitle{MOFDiff: Coarse-grained Diffusion for Metal-Organic Framework Design}. In \bibinfo{booktitle}{\emph{The Twelfth International Conference on Learning Representations}}.
\newblock


\bibitem[Gao et~al\mbox{.}(2025)]%
        {gao2025soft}
\bibfield{author}{\bibinfo{person}{Chang Gao}, \bibinfo{person}{Chujie Zheng}, \bibinfo{person}{Xiong-Hui Chen}, \bibinfo{person}{Kai Dang}, \bibinfo{person}{Shixuan Liu}, \bibinfo{person}{Bowen Yu}, \bibinfo{person}{An Yang}, \bibinfo{person}{Shuai Bai}, \bibinfo{person}{Jingren Zhou}, {and} \bibinfo{person}{Junyang Lin}.} \bibinfo{year}{2025}\natexlab{}.
\newblock \showarticletitle{Soft Adaptive Policy Optimization}.
\newblock \bibinfo{journal}{\emph{arXiv preprint arXiv:2511.20347}} (\bibinfo{year}{2025}).
\newblock


\bibitem[Gruver et~al\mbox{.}(2024)]%
        {gruverfine}
\bibfield{author}{\bibinfo{person}{Nate Gruver}, \bibinfo{person}{Anuroop Sriram}, \bibinfo{person}{Andrea Madotto}, \bibinfo{person}{Andrew~Gordon Wilson}, \bibinfo{person}{C~Lawrence Zitnick}, {and} \bibinfo{person}{Zachary~Ward Ulissi}.} \bibinfo{year}{2024}\natexlab{}.
\newblock \showarticletitle{Fine-Tuned Language Models Generate Stable Inorganic Materials as Text}. In \bibinfo{booktitle}{\emph{The Twelfth International Conference on Learning Representations}}.
\newblock


\bibitem[Guo et~al\mbox{.}(2025)]%
        {guo2025deepseek}
\bibfield{author}{\bibinfo{person}{Daya Guo}, \bibinfo{person}{Dejian Yang}, \bibinfo{person}{Haowei Zhang}, \bibinfo{person}{Junxiao Song}, \bibinfo{person}{Ruoyu Zhang}, \bibinfo{person}{Runxin Xu}, \bibinfo{person}{Qihao Zhu}, \bibinfo{person}{Shirong Ma}, \bibinfo{person}{Peiyi Wang}, \bibinfo{person}{Xiao Bi}, {et~al\mbox{.}}} \bibinfo{year}{2025}\natexlab{}.
\newblock \showarticletitle{Deepseek-r1: Incentivizing reasoning capability in llms via reinforcement learning}.
\newblock \bibinfo{journal}{\emph{arXiv preprint arXiv:2501.12948}} (\bibinfo{year}{2025}).
\newblock


\bibitem[Guo et~al\mbox{.}(2023)]%
        {guo2023can}
\bibfield{author}{\bibinfo{person}{Taicheng Guo}, \bibinfo{person}{Bozhao Nan}, \bibinfo{person}{Zhenwen Liang}, \bibinfo{person}{Zhichun Guo}, \bibinfo{person}{Nitesh Chawla}, \bibinfo{person}{Olaf Wiest}, \bibinfo{person}{Xiangliang Zhang}, {et~al\mbox{.}}} \bibinfo{year}{2023}\natexlab{}.
\newblock \showarticletitle{What can large language models do in chemistry? a comprehensive benchmark on eight tasks}.
\newblock \bibinfo{journal}{\emph{Advances in Neural Information Processing Systems}}  \bibinfo{volume}{36} (\bibinfo{year}{2023}), \bibinfo{pages}{59662--59688}.
\newblock


\bibitem[Horcajada et~al\mbox{.}(2010)]%
        {horcajada2010porous}
\bibfield{author}{\bibinfo{person}{Patricia Horcajada}, \bibinfo{person}{Tamim Chalati}, \bibinfo{person}{Christian Serre}, \bibinfo{person}{Brigitte Gillet}, \bibinfo{person}{Catherine Sebrie}, \bibinfo{person}{Tarek Baati}, \bibinfo{person}{Jarrod~F Eubank}, \bibinfo{person}{Daniela Heurtaux}, \bibinfo{person}{Pascal Clayette}, \bibinfo{person}{Christine Kreuz}, {et~al\mbox{.}}} \bibinfo{year}{2010}\natexlab{}.
\newblock \showarticletitle{Porous metal--organic-framework nanoscale carriers as a potential platform for drug delivery and imaging}.
\newblock \bibinfo{journal}{\emph{Nature materials}} \bibinfo{volume}{9}, \bibinfo{number}{2} (\bibinfo{year}{2010}), \bibinfo{pages}{172--178}.
\newblock


\bibitem[Hu et~al\mbox{.}(2022)]%
        {hu2022lora}
\bibfield{author}{\bibinfo{person}{Edward~J Hu}, \bibinfo{person}{Yelong Shen}, \bibinfo{person}{Phillip Wallis}, \bibinfo{person}{Zeyuan Allen-Zhu}, \bibinfo{person}{Yuanzhi Li}, \bibinfo{person}{Shean Wang}, \bibinfo{person}{Lu Wang}, \bibinfo{person}{Weizhu Chen}, {et~al\mbox{.}}} \bibinfo{year}{2022}\natexlab{}.
\newblock \showarticletitle{Lora: Low-rank adaptation of large language models.}
\newblock \bibinfo{journal}{\emph{ICLR}} \bibinfo{volume}{1}, \bibinfo{number}{2} (\bibinfo{year}{2022}), \bibinfo{pages}{3}.
\newblock


\bibitem[Inizan et~al\mbox{.}(2025)]%
        {inizan2025system}
\bibfield{author}{\bibinfo{person}{Theo~Jaffrelot Inizan}, \bibinfo{person}{Sherry Yang}, \bibinfo{person}{Aaron Kaplan}, \bibinfo{person}{Yen-hsu Lin}, \bibinfo{person}{Jian Yin}, \bibinfo{person}{Saber Mirzaei}, \bibinfo{person}{Mona Abdelgaid}, \bibinfo{person}{Ali~H Alawadhi}, \bibinfo{person}{KwangHwan Cho}, \bibinfo{person}{Zhiling Zheng}, {et~al\mbox{.}}} \bibinfo{year}{2025}\natexlab{}.
\newblock \showarticletitle{System of agentic AI for the discovery of metal-organic frameworks}.
\newblock \bibinfo{journal}{\emph{arXiv preprint arXiv:2504.14110}} (\bibinfo{year}{2025}).
\newblock


\bibitem[Jiao et~al\mbox{.}(2023)]%
        {jiao2023crystal}
\bibfield{author}{\bibinfo{person}{Rui Jiao}, \bibinfo{person}{Wenbing Huang}, \bibinfo{person}{Peijia Lin}, \bibinfo{person}{Jiaqi Han}, \bibinfo{person}{Pin Chen}, \bibinfo{person}{Yutong Lu}, {and} \bibinfo{person}{Yang Liu}.} \bibinfo{year}{2023}\natexlab{}.
\newblock \showarticletitle{Crystal structure prediction by joint equivariant diffusion}.
\newblock \bibinfo{journal}{\emph{Advances in Neural Information Processing Systems}}  \bibinfo{volume}{36} (\bibinfo{year}{2023}), \bibinfo{pages}{17464--17497}.
\newblock


\bibitem[Jiao et~al\mbox{.}(2024)]%
        {jiaospace}
\bibfield{author}{\bibinfo{person}{Rui Jiao}, \bibinfo{person}{Wenbing Huang}, \bibinfo{person}{Yu Liu}, \bibinfo{person}{Deli Zhao}, {and} \bibinfo{person}{Yang Liu}.} \bibinfo{year}{2024}\natexlab{}.
\newblock \showarticletitle{Space Group Constrained Crystal Generation}. In \bibinfo{booktitle}{\emph{The Twelfth International Conference on Learning Representations}}.
\newblock


\bibitem[Jiao et~al\mbox{.}(2025)]%
        {jiao2025mof}
\bibfield{author}{\bibinfo{person}{Rui Jiao}, \bibinfo{person}{Hanlin Wu}, \bibinfo{person}{Wenbing Huang}, \bibinfo{person}{Yuxuan Song}, \bibinfo{person}{Yawen Ouyang}, \bibinfo{person}{Yu Rong}, \bibinfo{person}{Tingyang Xu}, \bibinfo{person}{Pengju Wang}, \bibinfo{person}{Hao Zhou}, \bibinfo{person}{Wei-Ying Ma}, {et~al\mbox{.}}} \bibinfo{year}{2025}\natexlab{}.
\newblock \showarticletitle{Mof-bfn: Metal-organic frameworks structure prediction via bayesian flow networks}. In \bibinfo{booktitle}{\emph{The Thirty-ninth Annual Conference on Neural Information Processing Systems}}.
\newblock


\bibitem[Jin et~al\mbox{.}(2025)]%
        {jin2025mofchecker}
\bibfield{author}{\bibinfo{person}{Xin Jin}, \bibinfo{person}{Kevin~Maik Jablonka}, \bibinfo{person}{Elias Moubarak}, \bibinfo{person}{Yutao Li}, {and} \bibinfo{person}{Berend Smit}.} \bibinfo{year}{2025}\natexlab{}.
\newblock \showarticletitle{MOFChecker: a package for validating and correcting metal--organic framework (MOF) structures}.
\newblock \bibinfo{journal}{\emph{Digital Discovery}} \bibinfo{volume}{4}, \bibinfo{number}{6} (\bibinfo{year}{2025}), \bibinfo{pages}{1560--1569}.
\newblock


\bibitem[Kim et~al\mbox{.}(2017)]%
        {kim2017water}
\bibfield{author}{\bibinfo{person}{Hyunho Kim}, \bibinfo{person}{Sungwoo Yang}, \bibinfo{person}{Sameer~R Rao}, \bibinfo{person}{Shankar Narayanan}, \bibinfo{person}{Eugene~A Kapustin}, \bibinfo{person}{Hiroyasu Furukawa}, \bibinfo{person}{Ari~S Umans}, \bibinfo{person}{Omar~M Yaghi}, {and} \bibinfo{person}{Evelyn~N Wang}.} \bibinfo{year}{2017}\natexlab{}.
\newblock \showarticletitle{Water harvesting from air with metal-organic frameworks powered by natural sunlight}.
\newblock \bibinfo{journal}{\emph{Science}} \bibinfo{volume}{356}, \bibinfo{number}{6336} (\bibinfo{year}{2017}), \bibinfo{pages}{430--434}.
\newblock


\bibitem[Kim et~al\mbox{.}(2025)]%
        {kim2025flexible}
\bibfield{author}{\bibinfo{person}{Nayoung Kim}, \bibinfo{person}{Seongsu Kim}, {and} \bibinfo{person}{Sungsoo Ahn}.} \bibinfo{year}{2025}\natexlab{}.
\newblock \showarticletitle{Flexible MOF Generation with Torsion-Aware Flow Matching}.
\newblock \bibinfo{journal}{\emph{arXiv preprint arXiv:2505.17914}} (\bibinfo{year}{2025}).
\newblock


\bibitem[Kim et~al\mbox{.}(2024)]%
        {kimmofflow}
\bibfield{author}{\bibinfo{person}{Nayoung Kim}, \bibinfo{person}{Seongsu Kim}, \bibinfo{person}{Minsu Kim}, \bibinfo{person}{Jinkyoo Park}, {and} \bibinfo{person}{Sungsoo Ahn}.} \bibinfo{year}{2024}\natexlab{}.
\newblock \showarticletitle{MOFFlow: Flow Matching for Structure Prediction of Metal-Organic Frameworks}. In \bibinfo{booktitle}{\emph{The Thirteenth International Conference on Learning Representations}}.
\newblock


\bibitem[Kwon et~al\mbox{.}(2023)]%
        {kwon2023efficient}
\bibfield{author}{\bibinfo{person}{Woosuk Kwon}, \bibinfo{person}{Zhuohan Li}, \bibinfo{person}{Siyuan Zhuang}, \bibinfo{person}{Ying Sheng}, \bibinfo{person}{Lianmin Zheng}, \bibinfo{person}{Cody~Hao Yu}, \bibinfo{person}{Joseph~E. Gonzalez}, \bibinfo{person}{Hao Zhang}, {and} \bibinfo{person}{Ion Stoica}.} \bibinfo{year}{2023}\natexlab{}.
\newblock \showarticletitle{Efficient Memory Management for Large Language Model Serving with PagedAttention}. In \bibinfo{booktitle}{\emph{Proceedings of the ACM SIGOPS 29th Symposium on Operating Systems Principles}}.
\newblock


\bibitem[Lv et~al\mbox{.}(2025)]%
        {lv2025atomworld}
\bibfield{author}{\bibinfo{person}{Taoyuze Lv}, \bibinfo{person}{Alexander Chen}, \bibinfo{person}{Fengyu Xie}, \bibinfo{person}{Chu Wu}, \bibinfo{person}{Jeffrey Meng}, \bibinfo{person}{Dongzhan Zhou}, \bibinfo{person}{Bram Hoex}, \bibinfo{person}{Zhicheng Zhong}, {and} \bibinfo{person}{Tong Xie}.} \bibinfo{year}{2025}\natexlab{}.
\newblock \showarticletitle{AtomWorld: A Benchmark for Evaluating Spatial Reasoning in Large Language Models on Crystalline Materials}.
\newblock \bibinfo{journal}{\emph{arXiv preprint arXiv:2510.04704}} (\bibinfo{year}{2025}).
\newblock


\bibitem[McDonald et~al\mbox{.}(2015)]%
        {mcdonald2015cooperative}
\bibfield{author}{\bibinfo{person}{Thomas~M McDonald}, \bibinfo{person}{Jarad~A Mason}, \bibinfo{person}{Xueqian Kong}, \bibinfo{person}{Eric~D Bloch}, \bibinfo{person}{David Gygi}, \bibinfo{person}{Alessandro Dani}, \bibinfo{person}{Valentina Crocella}, \bibinfo{person}{Filippo Giordanino}, \bibinfo{person}{Samuel~O Odoh}, \bibinfo{person}{Walter~S Drisdell}, {et~al\mbox{.}}} \bibinfo{year}{2015}\natexlab{}.
\newblock \showarticletitle{Cooperative insertion of CO2 in diamine-appended metal-organic frameworks}.
\newblock \bibinfo{journal}{\emph{Nature}} \bibinfo{volume}{519}, \bibinfo{number}{7543} (\bibinfo{year}{2015}), \bibinfo{pages}{303--308}.
\newblock


\bibitem[Miller et~al\mbox{.}(2024)]%
        {miller2024flowmm}
\bibfield{author}{\bibinfo{person}{Benjamin~Kurt Miller}, \bibinfo{person}{Ricky~TQ Chen}, \bibinfo{person}{Anuroop Sriram}, {and} \bibinfo{person}{Brandon~M Wood}.} \bibinfo{year}{2024}\natexlab{}.
\newblock \showarticletitle{FlowMM: Generating Materials with Riemannian Flow Matching}. In \bibinfo{booktitle}{\emph{International Conference on Machine Learning}}. PMLR, \bibinfo{pages}{35664--35686}.
\newblock


\bibitem[Moosavi et~al\mbox{.}(2020)]%
        {moosavi2020understanding}
\bibfield{author}{\bibinfo{person}{Seyed~Mohamad Moosavi}, \bibinfo{person}{Aditya Nandy}, \bibinfo{person}{Kevin~Maik Jablonka}, \bibinfo{person}{Daniele Ongari}, \bibinfo{person}{Jon~Paul Janet}, \bibinfo{person}{Peter~G Boyd}, \bibinfo{person}{Yongjin Lee}, \bibinfo{person}{Berend Smit}, {and} \bibinfo{person}{Heather~J Kulik}.} \bibinfo{year}{2020}\natexlab{}.
\newblock \showarticletitle{Understanding the diversity of the metal-organic framework ecosystem}.
\newblock \bibinfo{journal}{\emph{Nature communications}} \bibinfo{volume}{11}, \bibinfo{number}{1} (\bibinfo{year}{2020}), \bibinfo{pages}{4068}.
\newblock


\bibitem[Neumann et~al\mbox{.}(2024)]%
        {neumann2024orbfastscalableneural}
\bibfield{author}{\bibinfo{person}{Mark Neumann}, \bibinfo{person}{James Gin}, \bibinfo{person}{Benjamin Rhodes}, \bibinfo{person}{Steven Bennett}, \bibinfo{person}{Zhiyi Li}, \bibinfo{person}{Hitarth Choubisa}, \bibinfo{person}{Arthur Hussey}, {and} \bibinfo{person}{Jonathan Godwin}.} \bibinfo{year}{2024}\natexlab{}.
\newblock \bibinfo{title}{Orb: A Fast, Scalable Neural Network Potential}.
\newblock
\showeprint[arxiv]{2410.22570}~[cond-mat.mtrl-sci]
\urldef\tempurl%
\url{https://arxiv.org/abs/2410.22570}
\showURL{%
\tempurl}


\bibitem[Ong et~al\mbox{.}(2013)]%
        {ong2013python}
\bibfield{author}{\bibinfo{person}{Shyue~Ping Ong}, \bibinfo{person}{William~Davidson Richards}, \bibinfo{person}{Anubhav Jain}, \bibinfo{person}{Geoffroy Hautier}, \bibinfo{person}{Michael Kocher}, \bibinfo{person}{Shreyas Cholia}, \bibinfo{person}{Dan Gunter}, \bibinfo{person}{Vincent~L Chevrier}, \bibinfo{person}{Kristin~A Persson}, {and} \bibinfo{person}{Gerbrand Ceder}.} \bibinfo{year}{2013}\natexlab{}.
\newblock \showarticletitle{Python Materials Genomics (pymatgen): A robust, open-source python library for materials analysis}.
\newblock \bibinfo{journal}{\emph{Computational Materials Science}}  \bibinfo{volume}{68} (\bibinfo{year}{2013}), \bibinfo{pages}{314--319}.
\newblock


\bibitem[Outreach(2025)]%
        {nobel2025chemistry}
\bibfield{author}{\bibinfo{person}{Nobel~Prize Outreach}.} \bibinfo{year}{2025}\natexlab{}.
\newblock \bibinfo{booktitle}{\emph{Nobel Prize in Chemistry 2025}}.
\newblock
\urldef\tempurl%
\url{https://www.nobelprize.org/prizes/chemistry/2025/summary/}
\showURL{%
\tempurl}


\bibitem[O’Boyle(2012)]%
        {o2012towards}
\bibfield{author}{\bibinfo{person}{Noel~M O’Boyle}.} \bibinfo{year}{2012}\natexlab{}.
\newblock \showarticletitle{Towards a Universal SMILES representation-A standard method to generate canonical SMILES based on the InChI}.
\newblock \bibinfo{journal}{\emph{Journal of cheminformatics}} \bibinfo{volume}{4}, \bibinfo{number}{1} (\bibinfo{year}{2012}), \bibinfo{pages}{22}.
\newblock


\bibitem[Pun et~al\mbox{.}(2025)]%
        {pun2025generating}
\bibfield{author}{\bibinfo{person}{Ava Pun}, \bibinfo{person}{Kangle Deng}, \bibinfo{person}{Ruixuan Liu}, \bibinfo{person}{Deva Ramanan}, \bibinfo{person}{Changliu Liu}, {and} \bibinfo{person}{Jun-Yan Zhu}.} \bibinfo{year}{2025}\natexlab{}.
\newblock \showarticletitle{Generating physically stable and buildable brick structures from text}. In \bibinfo{booktitle}{\emph{Proceedings of the IEEE/CVF International Conference on Computer Vision}}. \bibinfo{pages}{14798--14809}.
\newblock


\bibitem[Qwen et~al\mbox{.}(2025)]%
        {qwen2025qwen25technicalreport}
\bibfield{author}{\bibinfo{person}{Qwen}, \bibinfo{person}{:}, \bibinfo{person}{An Yang}, \bibinfo{person}{Baosong Yang}, \bibinfo{person}{Beichen Zhang}, \bibinfo{person}{Binyuan Hui}, \bibinfo{person}{Bo Zheng}, \bibinfo{person}{Bowen Yu}, \bibinfo{person}{Chengyuan Li}, \bibinfo{person}{Dayiheng Liu}, \bibinfo{person}{Fei Huang}, \bibinfo{person}{Haoran Wei}, \bibinfo{person}{Huan Lin}, \bibinfo{person}{Jian Yang}, \bibinfo{person}{Jianhong Tu}, \bibinfo{person}{Jianwei Zhang}, \bibinfo{person}{Jianxin Yang}, \bibinfo{person}{Jiaxi Yang}, \bibinfo{person}{Jingren Zhou}, \bibinfo{person}{Junyang Lin}, \bibinfo{person}{Kai Dang}, \bibinfo{person}{Keming Lu}, \bibinfo{person}{Keqin Bao}, \bibinfo{person}{Kexin Yang}, \bibinfo{person}{Le Yu}, \bibinfo{person}{Mei Li}, \bibinfo{person}{Mingfeng Xue}, \bibinfo{person}{Pei Zhang}, \bibinfo{person}{Qin Zhu}, \bibinfo{person}{Rui Men}, \bibinfo{person}{Runji Lin}, \bibinfo{person}{Tianhao Li}, \bibinfo{person}{Tianyi Tang}, \bibinfo{person}{Tingyu Xia},
  \bibinfo{person}{Xingzhang Ren}, \bibinfo{person}{Xuancheng Ren}, \bibinfo{person}{Yang Fan}, \bibinfo{person}{Yang Su}, \bibinfo{person}{Yichang Zhang}, \bibinfo{person}{Yu Wan}, \bibinfo{person}{Yuqiong Liu}, \bibinfo{person}{Zeyu Cui}, \bibinfo{person}{Zhenru Zhang}, {and} \bibinfo{person}{Zihan Qiu}.} \bibinfo{year}{2025}\natexlab{}.
\newblock \bibinfo{title}{Qwen2.5 Technical Report}.
\newblock
\showeprint[arxiv]{2412.15115}~[cs.CL]
\urldef\tempurl%
\url{https://arxiv.org/abs/2412.15115}
\showURL{%
\tempurl}


\bibitem[Rasley et~al\mbox{.}(2020)]%
        {rasley2020deepspeed}
\bibfield{author}{\bibinfo{person}{Jeff Rasley}, \bibinfo{person}{Samyam Rajbhandari}, \bibinfo{person}{Olatunji Ruwase}, {and} \bibinfo{person}{Yuxiong He}.} \bibinfo{year}{2020}\natexlab{}.
\newblock \showarticletitle{Deepspeed: System optimizations enable training deep learning models with over 100 billion parameters}. In \bibinfo{booktitle}{\emph{Proceedings of the 26th ACM SIGKDD international conference on knowledge discovery \& data mining}}. \bibinfo{pages}{3505--3506}.
\newblock


\bibitem[Rhodes et~al\mbox{.}(2025)]%
        {rhodes2025orbv3atomisticsimulationscale}
\bibfield{author}{\bibinfo{person}{Benjamin Rhodes}, \bibinfo{person}{Sander Vandenhaute}, \bibinfo{person}{Vaidotas Šimkus}, \bibinfo{person}{James Gin}, \bibinfo{person}{Jonathan Godwin}, \bibinfo{person}{Tim Duignan}, {and} \bibinfo{person}{Mark Neumann}.} \bibinfo{year}{2025}\natexlab{}.
\newblock \bibinfo{title}{Orb-v3: atomistic simulation at scale}.
\newblock
\showeprint[arxiv]{2504.06231}~[cond-mat.mtrl-sci]
\urldef\tempurl%
\url{https://arxiv.org/abs/2504.06231}
\showURL{%
\tempurl}


\bibitem[Shao et~al\mbox{.}(2024)]%
        {shao2024deepseekmath}
\bibfield{author}{\bibinfo{person}{Zhihong Shao}, \bibinfo{person}{Peiyi Wang}, \bibinfo{person}{Qihao Zhu}, \bibinfo{person}{Runxin Xu}, \bibinfo{person}{Junxiao Song}, \bibinfo{person}{Xiao Bi}, \bibinfo{person}{Haowei Zhang}, \bibinfo{person}{Mingchuan Zhang}, \bibinfo{person}{YK Li}, {et~al\mbox{.}}} \bibinfo{year}{2024}\natexlab{}.
\newblock \showarticletitle{Deepseekmath: Pushing the limits of mathematical reasoning in open language models}.
\newblock \bibinfo{journal}{\emph{arXiv preprint arXiv:2402.03300}} (\bibinfo{year}{2024}).
\newblock


\bibitem[Shen et~al\mbox{.}(2023)]%
        {shen2023large}
\bibfield{author}{\bibinfo{person}{Tianhao Shen}, \bibinfo{person}{Renren Jin}, \bibinfo{person}{Yufei Huang}, \bibinfo{person}{Chuang Liu}, \bibinfo{person}{Weilong Dong}, \bibinfo{person}{Zishan Guo}, \bibinfo{person}{Xinwei Wu}, \bibinfo{person}{Yan Liu}, {and} \bibinfo{person}{Deyi Xiong}.} \bibinfo{year}{2023}\natexlab{}.
\newblock \showarticletitle{Large language model alignment: A survey}.
\newblock \bibinfo{journal}{\emph{arXiv preprint arXiv:2309.15025}} (\bibinfo{year}{2023}).
\newblock


\bibitem[Sriram et~al\mbox{.}(2024)]%
        {sriram2024flowllm}
\bibfield{author}{\bibinfo{person}{Anuroop Sriram}, \bibinfo{person}{Benjamin~K Miller}, \bibinfo{person}{Ricky~T Chen}, {and} \bibinfo{person}{Brandon Wood}.} \bibinfo{year}{2024}\natexlab{}.
\newblock \showarticletitle{Flowllm: Flow matching for material generation with large language models as base distributions}.
\newblock \bibinfo{journal}{\emph{Advances in Neural Information Processing Systems}}  \bibinfo{volume}{37} (\bibinfo{year}{2024}), \bibinfo{pages}{46025--46046}.
\newblock


\bibitem[Wang et~al\mbox{.}(2025)]%
        {wang2025crystalicl}
\bibfield{author}{\bibinfo{person}{Ruobing Wang}, \bibinfo{person}{Qiaoyu Tan}, \bibinfo{person}{Yili Wang}, \bibinfo{person}{Ying Wang}, {and} \bibinfo{person}{Xin Wang}.} \bibinfo{year}{2025}\natexlab{}.
\newblock \showarticletitle{CrystalICL: Enabling In-Context Learning for Crystal Generation}. In \bibinfo{booktitle}{\emph{Proceedings of the 2025 Conference on Empirical Methods in Natural Language Processing}}. \bibinfo{pages}{18440--18455}.
\newblock


\bibitem[Wang et~al\mbox{.}(2024)]%
        {wang2024llama}
\bibfield{author}{\bibinfo{person}{Zhengyi Wang}, \bibinfo{person}{Jonathan Lorraine}, \bibinfo{person}{Yikai Wang}, \bibinfo{person}{Hang Su}, \bibinfo{person}{Jun Zhu}, \bibinfo{person}{Sanja Fidler}, {and} \bibinfo{person}{Xiaohui Zeng}.} \bibinfo{year}{2024}\natexlab{}.
\newblock \showarticletitle{Llama-mesh: Unifying 3d mesh generation with language models}.
\newblock \bibinfo{journal}{\emph{arXiv preprint arXiv:2411.09595}} (\bibinfo{year}{2024}).
\newblock


\bibitem[Weininger et~al\mbox{.}(1989)]%
        {weininger1989smiles}
\bibfield{author}{\bibinfo{person}{David Weininger}, \bibinfo{person}{Arthur Weininger}, {and} \bibinfo{person}{Joseph~L Weininger}.} \bibinfo{year}{1989}\natexlab{}.
\newblock \showarticletitle{SMILES. 2. Algorithm for generation of unique SMILES notation}.
\newblock \bibinfo{journal}{\emph{Journal of chemical information and computer sciences}} \bibinfo{volume}{29}, \bibinfo{number}{2} (\bibinfo{year}{1989}), \bibinfo{pages}{97--101}.
\newblock


\bibitem[Willems et~al\mbox{.}(2012)]%
        {willems2012algorithms}
\bibfield{author}{\bibinfo{person}{Thomas~F Willems}, \bibinfo{person}{Chris~H Rycroft}, \bibinfo{person}{Michaeel Kazi}, \bibinfo{person}{Juan~C Meza}, {and} \bibinfo{person}{Maciej Haranczyk}.} \bibinfo{year}{2012}\natexlab{}.
\newblock \showarticletitle{Algorithms and tools for high-throughput geometry-based analysis of crystalline porous materials}.
\newblock \bibinfo{journal}{\emph{Microporous and Mesoporous Materials}} \bibinfo{volume}{149}, \bibinfo{number}{1} (\bibinfo{year}{2012}), \bibinfo{pages}{134--141}.
\newblock


\bibitem[Wu et~al\mbox{.}(2025)]%
        {wu2025periodic}
\bibfield{author}{\bibinfo{person}{Hanlin Wu}, \bibinfo{person}{Yuxuan Song}, \bibinfo{person}{Jingjing Gong}, \bibinfo{person}{Ziyao Cao}, \bibinfo{person}{Yawen Ouyang}, \bibinfo{person}{Jianbing Zhang}, \bibinfo{person}{Hao Zhou}, \bibinfo{person}{Wei-Ying Ma}, {and} \bibinfo{person}{Jingjing Liu}.} \bibinfo{year}{2025}\natexlab{}.
\newblock \showarticletitle{A Periodic Bayesian Flow for Material Generation}. In \bibinfo{booktitle}{\emph{ICLR}}.
\newblock


\bibitem[Wu et~al\mbox{.}(2024)]%
        {wu2024leveraging}
\bibfield{author}{\bibinfo{person}{Zhenxing Wu}, \bibinfo{person}{Odin Zhang}, \bibinfo{person}{Xiaorui Wang}, \bibinfo{person}{Li Fu}, \bibinfo{person}{Huifeng Zhao}, \bibinfo{person}{Jike Wang}, \bibinfo{person}{Hongyan Du}, \bibinfo{person}{Dejun Jiang}, \bibinfo{person}{Yafeng Deng}, \bibinfo{person}{Dongsheng Cao}, {et~al\mbox{.}}} \bibinfo{year}{2024}\natexlab{}.
\newblock \showarticletitle{Leveraging language model for advanced multiproperty molecular optimization via prompt engineering}.
\newblock \bibinfo{journal}{\emph{Nature Machine Intelligence}} \bibinfo{volume}{6}, \bibinfo{number}{11} (\bibinfo{year}{2024}), \bibinfo{pages}{1359--1369}.
\newblock


\bibitem[Xu et~al\mbox{.}(2025)]%
        {xu2025plaid++}
\bibfield{author}{\bibinfo{person}{Andy Xu}, \bibinfo{person}{Rohan Desai}, \bibinfo{person}{Larry Wang}, \bibinfo{person}{Gabriel Hope}, {and} \bibinfo{person}{Ethan Ritz}.} \bibinfo{year}{2025}\natexlab{}.
\newblock \showarticletitle{Plaid++: A preference aligned language model for targeted inorganic materials design}.
\newblock \bibinfo{journal}{\emph{arXiv preprint arXiv:2509.07150}} (\bibinfo{year}{2025}).
\newblock


\bibitem[Yang et~al\mbox{.}(2025)]%
        {yang2025qwen3}
\bibfield{author}{\bibinfo{person}{An Yang}, \bibinfo{person}{Anfeng Li}, \bibinfo{person}{Baosong Yang}, \bibinfo{person}{Beichen Zhang}, \bibinfo{person}{Binyuan Hui}, \bibinfo{person}{Bo Zheng}, \bibinfo{person}{Bowen Yu}, \bibinfo{person}{Chang Gao}, \bibinfo{person}{Chengen Huang}, \bibinfo{person}{Chenxu Lv}, {et~al\mbox{.}}} \bibinfo{year}{2025}\natexlab{}.
\newblock \showarticletitle{Qwen3 technical report}.
\newblock \bibinfo{journal}{\emph{arXiv preprint arXiv:2505.09388}} (\bibinfo{year}{2025}).
\newblock


\bibitem[Zhao et~al\mbox{.}(2024)]%
        {zhao2024swiftascalablelightweightinfrastructure}
\bibfield{author}{\bibinfo{person}{Yuze Zhao}, \bibinfo{person}{Jintao Huang}, \bibinfo{person}{Jinghan Hu}, \bibinfo{person}{Xingjun Wang}, \bibinfo{person}{Yunlin Mao}, \bibinfo{person}{Daoze Zhang}, \bibinfo{person}{Zeyinzi Jiang}, \bibinfo{person}{Zhikai Wu}, \bibinfo{person}{Baole Ai}, \bibinfo{person}{Ang Wang}, \bibinfo{person}{Wenmeng Zhou}, {and} \bibinfo{person}{Yingda Chen}.} \bibinfo{year}{2024}\natexlab{}.
\newblock \bibinfo{title}{SWIFT:A Scalable lightWeight Infrastructure for Fine-Tuning}.
\newblock
\showeprint[arxiv]{2408.05517}~[cs.CL]
\urldef\tempurl%
\url{https://arxiv.org/abs/2408.05517}
\showURL{%
\tempurl}


\bibitem[Zheng et~al\mbox{.}(2024a)]%
        {zheng2024sglang}
\bibfield{author}{\bibinfo{person}{Lianmin Zheng}, \bibinfo{person}{Liangsheng Yin}, \bibinfo{person}{Zhiqiang Xie}, \bibinfo{person}{Chuyue~Livia Sun}, \bibinfo{person}{Jeff Huang}, \bibinfo{person}{Cody~Hao Yu}, \bibinfo{person}{Shiyi Cao}, \bibinfo{person}{Christos Kozyrakis}, \bibinfo{person}{Ion Stoica}, \bibinfo{person}{Joseph~E Gonzalez}, {et~al\mbox{.}}} \bibinfo{year}{2024}\natexlab{a}.
\newblock \showarticletitle{Sglang: Efficient execution of structured language model programs}.
\newblock \bibinfo{journal}{\emph{Advances in neural information processing systems}}  \bibinfo{volume}{37} (\bibinfo{year}{2024}), \bibinfo{pages}{62557--62583}.
\newblock


\bibitem[Zheng et~al\mbox{.}(2024b)]%
        {zheng2024llamafactory}
\bibfield{author}{\bibinfo{person}{Yaowei Zheng}, \bibinfo{person}{Richong Zhang}, \bibinfo{person}{Junhao Zhang}, \bibinfo{person}{Yanhan Ye}, \bibinfo{person}{Zheyan Luo}, \bibinfo{person}{Zhangchi Feng}, {and} \bibinfo{person}{Yongqiang Ma}.} \bibinfo{year}{2024}\natexlab{b}.
\newblock \showarticletitle{Llamafactory: Unified efficient fine-tuning of 100+ language models}.
\newblock \bibinfo{journal}{\emph{arXiv preprint arXiv:2403.13372}} (\bibinfo{year}{2024}).
\newblock


\end{thebibliography}

\appendix

\section{Appendix}
\subsection{Training Details}
\subsubsection{Spatial-aware Continual Pre-training}\label{app:train_cpt}
We utilize Qwen-3 8B~\citep{yang2025qwen3} as our backbone model. To adapt the general-purpose model to the target domain, we perform full-parameter continual pre-training (CPT) using the \textsc{LLaMA-Factory} framework~\citep{zheng2024llamafactory}. The experiment is conducted on 4 $\times$ NVIDIA H800-80GB GPUs.

\paragraph{Data Processing and Efficiency.}
To maximize computational efficiency and training throughput, we employ the sequence packing strategy. Instead of padding individual samples to the maximum length, multiple text segments from the dataset are concatenated to fill the context window of 3,072 tokens.

\paragraph{Optimization Setup.} 
To accelerate training and reduce memory footprint, we employ DeepSpeed ZeRO Stage 3~\citep{rasley2020deepspeed} alongside Flash Attention 2~\citep{dao2023flashattention}. The training is conducted using bfloat16 (BF16) precision to ensure numerical stability and efficiency. We optimize the model using the AdamW optimizer with a standard auto-regressive language modeling objective.

\paragraph{Hyperparameters.} 
The learning rate is configured with a peak value of 5e-5, following a cosine decay schedule with a 5\% warmup period. The model is trained for 3 epochs with a global batch size of 128.

\subsubsection{Structural Supervised Fine-tuning}
Initialized with the last checkpoint from the CPT stage, we perform full-parameter Supervised Fine-Tuning (SFT) using the \textsc{LLaMA-Factory} framework. The training is conducted on 5 $\times$ NVIDIA H800-80GB GPUs.

\paragraph{Optimization Setup.}
We maintain the same optimization configuration as in \ref{app:train_cpt}.

\paragraph{Hyperparameters.}
We set the maximum input length to 2048 tokens, which covers 99\% of the data. The learning rate is configured with a peak value of 2e-5, following a cosine decay schedule with a 3\% warmup period. The model was trained for 12,000 steps with a global batch size of 100.

\subsubsection{Matching-driven Reinforcement Learning}\label{app:rl}
We utilize the SFT checkpoint as the initialization for the policy model to conduct the matching-driven RL using the \textsc{ms-swift} framework~\cite{zhao2024swiftascalablelightweightinfrastructure}. The training is performed on 8 $\times$ NVIDIA H800-80GB GPUs.

\paragraph{Reward Calculation.} Instead of explicitly calculating the minimum site tolerance $\tau$, we hierarchically verify whether the condition in \Cref{eq:reward} is met. The core Python implementation for the reward calculation is as follows:
% \begin{minted}[
%     linenos,
%     frame=lines,
%     framesep=2mm,
%     fontsize=\small,
%     breaklines,
%     breakanywhere,
%     breakautoindent,
%     autogobble
% ]{python}
\begin{lstlisting}[
    language=Python,
    numbers=left,
    frame=lines,
    framesep=2mm,
    basicstyle=\small\ttfamily,
    keywordstyle=\color{blue},
    commentstyle=\color{codegreen},
    stringstyle=\color{codepurple},
    numberstyle=\tiny\color{codegray},
    backgroundcolor=\color{backcolour},
    breaklines=true,
    breakatwhitespace=false,
    columns=fullflexible,
    keepspaces=true,
    showstringspaces=false,
    tabsize=4,
    xleftmargin=2em,
    framexleftmargin=1.5em,
    emph={
        self,True,False,None,
        int,float,str,list,dict,set,tuple,
        len,range,enumerate,zip,print,
        torch,numpy,np,pandas,pd
    },
    emphstyle=\color{teal}
]
from pymatgen.core import Structure
from pymatgen.analysis.structure_matcher import StructureMatcher

def reward(S_i: Structure, S_gt: Structure):
    # Assume S_i has been successfully parsed
    
    # Initialize matchers with increasing site tolerances (stol)
    matcher_1 = StructureMatcher(stol=0.5, ltol=0.3, angle_tol=10)
    matcher_2 = StructureMatcher(stol=0.75, ltol=0.3, angle_tol=10)
    matcher_3 = StructureMatcher(stol=1.0, ltol=0.3, angle_tol=10)
    
    # Hierarchical check
    rmsd = matcher_1.get_rms_dist(S_i, S_gt)
    if rmsd is not None: # Case: \tau <= 0.5
        return 1.0 + 0.5 * math.exp(-4 * rmsd[0])
    elif matcher_2.get_rms_dist(S_i, S_gt) is not None: # Case: 0.5 < \tau <= 0.75 
        return 0.6
    elif matcher_3.get_rms_dist(S_i, S_gt) is not None: # Case: 0.75 < \tau <= 1.0
        return 0.3
    else:
        return 0.0
%\end{minted}
\end{lstlisting}

\paragraph{Generation and Rollout.} 
To ensure efficient exploration, we utilize the \texttt{vLLM} engine~\citep{kwon2023efficient} for high-throughput sequence generation. For each prompt, the model samples a group of 8 outputs. The sampling is performed with a temperature of $0.9$, top-$p$ of $0.99$, and top-$k$ of $50$. We enable dynamic sampling with a maximum of 2 resample attempts to handle generation failures. The maximum completion length is restricted to 1,600 tokens.

\paragraph{Optimization Setup.}
We maintained the same optimization configuration as in \ref{app:train_cpt}

\paragraph{Hyperparameters.} 
The two SAPO temperate parameters are set to $\tau_{pos}=1.0, \tau_{neg}=1.05$.
The learning rate is configured with a peak value of 1e-6, following a cosine decay schedule with a 5\% warmup period. The model is trained for 2 epochs with a per-device batch size of 16.

\subsection{Sampling Details}
We sample MOF structures by \texttt{vLLM} on a single NVIDIA A100-80 GB GPU to ensure a fair comparison of sampling efficiency with the baselines (see the following section). 
For the one-sample evaluation setting, the sampling temperature, top-$p$, and top-$k$ are set to 0.1, 0.6, and 10, respectively, encouraging low-variance and deterministic generation. 
For the five-sample evaluation, these parameters are adjusted to a temperature of 0.9, top-$p$ of 0.99, and top-$k$ of 50 to promote greater output diversity.

\subsection{Implementation Details of Baselines}\label{app:baseline}
The match rate and RMSE values for the denoising-based baselines are taken directly from \citet{kimmofflow} and \citet{jiao2025mof}. 
To enable a fair comparison of sampling efficiency, we run MOFFlow~\citep{kimmofflow} and MOF-BFN~\citep{jiao2025mof} on a single NVIDIA A100-80 GB GPU to sample MOF structures. 
Specifically, we adopt the \texttt{Batch} version of MOFFlow, which offers an accelerated implementation. The runtime of DiffCSP~\citep{jiao2023crystal} is directly taken from \citet{kimmofflow}.

For PLaID++~\citep{xu2025plaid++}, we reimplement it by fine-tuning a Qwen-3 8B model using Wyckoff-based textual encodings of MOFs. The prompt template is as follows.
\begin{tcolorbox}[
colback=white, 
colframe=black!20, 
fontupper=\ttfamily\small, 
halign=justify,
left=1.5mm,
right=0mm
] 
\textbf{Instruction:} Below is a description of a crystal. The chemical formula is \textcolor{blue}{[Chemical Formula]}. The spacegroup number is \textcolor{blue}{[Spacegroup]}. Generate a description of the lengths and angles of the lattice vectors and then the element type and coordinates for each atom within the lattice.

\textbf{Response:} \textcolor{red}{[Wykoff-based String]}
\end{tcolorbox}
\noindent We exclude the preference alignment stage in PLaID++ because it relies on external evaluation tools, which would result in an unfair comparison.
During training, the maximum input length is set to 8192 tokens, corresponding to approximately the 90th percentile of the sequence length distribution, to balance coverage and computational efficiency.
As the long prompts make full-parameter fine-tuning prohibitively expensive, we adopt LoRA~\citep{hu2022lora} with default settings (rank=8, $\alpha$=32) used by \citet{xu2025plaid++}.
The learning rate follows a cosine decay schedule with a peak value of 1e-5 and a 3\% warmup period. The model is trained for 5,000 steps with a global batch size of 160.

During evaluation, we observe that almost all generated Wyckoff strings cannot be parsed into valid structures, likely because the model struggles to produce legal Wyckoff positions for a large number of atoms. Accordingly, we adopt a lenient evaluation setting that ignores Wyckoff positions and matches structures solely based on the generated atomic species and coordinates.

\subsection{Construction of Local Coordinates for Building Blocks}\label{app:local_coord}
To ensure that local coordinates remain consistent regardless of the building block's initial global pose (\emph{SE(3)}-invariance), we adopt the global-to-local transformation scheme introduced by \citet{kimmofflow}. 
This process first enforces \emph{translational invariance} by subtracting the block geometric centroid to center the atomic coordinates. \emph{Rotational invariance} is then achieved by aligning the building block with a deterministic local frame derived from Principal Component Analysis (PCA) axes, with sign ambiguity resolved using a mass-weighted reference vector. 
In addition, unit cell canonicalization via Niggli reduction is applied prior to this transformation to ensure consistency across periodic representations. Details of this procedure can be found in \citet{kimmofflow}.

\subsection{Rotation Representation}\label{app:rotation}
\subsubsection{Euler Angle Conversion}
Given the rotation matrix
\begin{equation}
\mathbf{R}=\left[\begin{array}{lll}
r_{11} & r_{12} & r_{13} \\
r_{21} & r_{22} & r_{23} \\
r_{31} & r_{32} & r_{33}
\end{array}\right],
\end{equation}
the Euler angles are calculated as:
\begin{equation}
\begin{aligned}
    &\omega=\mathrm{arcsin}(-r_{31}),\qquad\omega\in[-\pi/2,\pi/2]\\
    &\phi=\mathrm{arctan2}(r_{32},r_{33}),\quad\phi\in[-\pi,\pi]\\
    &\psi=\mathrm{arctan2}(r_{21},r_{11}),\quad\psi\in[-\pi,\pi]
\end{aligned}
\end{equation}
where $\mathrm{arctan2}(y,x)$ is the two-argument arctangent function which handles the quadrant of the angle.

\subsubsection{Axis-angle Rotation Representation}
In the axis–angle formulation, a 3D rotation is represented by a rotation vector obtained by multiplying a unit rotation axis $\mathbf{a} = (a_x, a_y, a_z)$ with a rotation angle $\psi$. Specifically, the rotation vector $\mathbf{m} \in \mathbb{R}^3$ is defined as
\begin{equation}
\mathbf{m} = (m_x, m_y, m_z) = \psi\mathbf{a} = (\psi a_x, \psi a_y, \psi a_z).
\end{equation}
Specifically, the rotation angle is obtained by
\begin{equation}
    \psi=\cos ^{-1}\left(\frac{\operatorname{tr}\left(\mathbf{R}\right)-1}{2}\right),
\end{equation}
and the rotation axis is calculated by
\begin{equation}
\begin{aligned}
a_{x} & =\left(r_{32}-r_{23}\right) / 2 \sin \psi, \\
a_{y} & =\left(r_{13}-r_{31}\right) / 2 \sin \psi, \\
a_{z} & =\left(r_{21}-r_{12}\right) / 2 \sin \psi .
\end{aligned}
\end{equation}

\subsection{Prompt Details}\label{app:prompt}
The detailed prompt template for spatial-aware CPT is provided as follows:

\begin{tcolorbox}[
  colback=white,
  colframe=black!20,
  fontupper=\ttfamily\footnotesize,
  halign=justify,
  left=1mm,
  right=0mm
]
The 3D crystal structure of a metal-organic framework (MOF) is described. MOFs represent a class of highly ordered porous materials that assemble modular architectures through metal node and organic linker connections. This metal-organic framework adopts the \textcolor{blue}{[Topo Code]} topology, with the topology code defined as the standard symbol in the Reticular Chemistry Structure Resource (RCSR) database. \textcolor{blue}{[Topo Description]}. The periodic lattice is specified by parameters: a = \textcolor{blue}{[a]} Å, b = \textcolor{blue}{[b]} Å, c = \textcolor{blue}{[c]} Å, and angles $\alpha$ = \textcolor{blue}{[$\alpha$]}°, $\beta$ = \textcolor{blue}{[$\beta$]}°, $\gamma$ = \textcolor{blue}{[$\gamma$]}°. 

Each building block is treated as a rigid entity with a standardized local coordinate system. 

In the local frame, the main structural PCA axis is aligned with [1, 0, 0]. The local geometry of a block is encapsulated by the spatial extent of atoms along the primary axes in Å. Local properties of building blocks are provided: 

[0] SMILES=\textcolor{blue}{[SMILES]}, molecular weight=\textcolor{blue}{[Molecular Weight]}, PCA span=\textcolor{blue}{[PCA Span]} 

[1] ... 

Global placement for each building block is defined in the lattice coordinate system, covering translation vectors, rotation angles, and the rotated main axis. The translation vector ([tx, ty, tz]) denotes the block's center position in the unit cell as fractional coordinates. Rotation angles ([roll pitch yaw]), expressed in radians via Euler angles, describe the block's orientation and the local main axis after rotation. Additionally, per block, the rotated main axis indicates the local main axis orientation following rotation. Specifics for each building block follow: 

[0] Translation=\textcolor{blue}{[Translation]}, rotation=\textcolor{blue}{[Rotation]}, rotated main axis=\textcolor{blue}{[Rotated Principal Axis]} 

[1] ..
\end{tcolorbox}

\noindent To enhance prompt diversity, we rewrite the base template to generate 15 distinct variants using DeepSeek~\citep{deepseekai2024deepseekv3technicalreport}.

The detailed SFT prompt template is provided as follows:
\begin{tcolorbox}[
  colback=white,
  colframe=black!20,
  fontupper=\ttfamily\footnotesize,
  halign=justify,
  left=1mm,
  right=0mm
]
\textbf{Instruction:} You are tasked with predicting the 3D crystal structure of a Metal-Organic Framework (MOF).

MOFs are highly ordered porous materials formed by connecting metal-containing nodes with
organic linkers, creating modular structures.

Given the SMILES of all MOF building blocks, predict the complete 3D crystal structure
configuration including:

Lattice parameters in the format 'a b c $\alpha$ $\beta$ $\gamma$' (where a, b, c are unit cell lengths, and $\alpha$, $\beta$,
$\gamma$ are unit cell angles).

For each building block (maintaining the exact same order as provided in the input):

Translation vector ([tx ty tz]): The position of the building block's center within the unit
cell, expressed in fractional coordinates.

Rotation angles ([roll pitch yaw]): The orientation of the building block, represented in
radians using Euler angles.

Output Format:

First line: Lattice parameters, i.e.: a b c $\alpha$ $\beta$ $\gamma$.

Subsequent lines (one per building block): [k] tx ty tz roll pitch yaw

k: The 0-based index of the building block.

tx ty tz: Fractional coordinates within the unit cell.

roll pitch yaw: Euler angles in radians.

\textcolor{blue}{[SMILES]} \textcolor{blue}{[SMILES]} ...

\textbf{Response:} \textcolor{red}{a b c $\alpha$ $\beta$ $\gamma$}

\textcolor{red}{[0] tx ty tz roll pitch yaw} 

\textcolor{red}{[1] tx ty tz roll pitch yaw}

...
\end{tcolorbox}

\end{document}